\documentclass[iicol,sn-basic,pdflatex]{sn-jnl}

\usepackage{graphicx}%
\usepackage{multirow}%
\usepackage{amsmath,amssymb,amsfonts}%
\usepackage{dsfont}
\usepackage{amsthm}%
\usepackage{mathrsfs}%
\usepackage[title]{appendix}%
\usepackage{xcolor}%
\usepackage{textcomp}%
\usepackage{manyfoot}%
\usepackage{booktabs}%
\usepackage{algorithmic}
\usepackage{algorithm}%
\usepackage{listings}%
\usepackage{subcaption}
\usepackage{makecell}  
\usepackage{array}     
\usepackage{caption}
\usepackage{threeparttable}

\theoremstyle{thmstyleone}%
\newtheorem{theorem}{Theorem}
\newtheorem{proposition}[theorem]{Proposition}%

\theoremstyle{thmstyletwo}%

\theoremstyle{thmstylethree}%

\begin{document}

\title[Article Title]{Hyperbolic Dual Feature Augmentation for Open-Environment}






\author[1]{Peilin Yu}\email{peilinyu@bit.edu.cn}

\author[1,2]{Yuwei Wu}\email{wuyuwei@bit.edu.cn}

\author*[1]{Zhi Gao}\email{zhi.gao@bit.edu.cn}
\author[1]{Xiaomeng Fan}\email{fanxiaomeng@bit.edu.cn} 

\author*[2]{Shuo Yang}\email{yangshuo@smbu.edu.cn}

\author[2,1]{Yunde Jia}\email{jiayunde@bit.edu.cn}

\affil[1]{\orgdiv{Beijing Laboratory of Intelligent
Information Technology}, \orgname{School of Computer Science, Beijing
Institute of Technology (BIT)}, \orgaddress{ \city{Beijing}, \postcode{100081}, \country{P.R. China}}}

\affil[2]{\orgdiv{Guangdong Laboratory of Machine
Perception and Intelligent Computing}, \\ \orgname{Shenzhen MSU-BIT University}, \orgaddress{ \city{Shenzhen}, \postcode{518172}, \country{P.R. China}}}


\abstract{
Feature augmentation generates novel samples in the feature space, providing an effective way to enhance the generalization ability of learning algorithms with hyperbolic geometry. Most hyperbolic feature augmentation is confined to closed-environment, assuming the number of classes is fixed (\emph{i.e.}, seen classes) and generating features only for these classes. 
In this paper, we propose a hyperbolic dual feature augmentation method for open-environment, which augments features for both seen and unseen classes in the hyperbolic space. To obtain a more precise approximation of the real data distribution for efficient training, (1) we adopt a neural ordinary differential equation module, enhanced by meta-learning, estimating the feature distributions of both seen and unseen classes; (2) we then introduce a regularizer to preserve the latent hierarchical structures of data in the hyperbolic space; (3) we also derive an upper bound for the hyperbolic dual augmentation loss, allowing us to train a hyperbolic model using infinite augmentations for seen and unseen classes. 
Extensive experiments on five open-environment tasks: class-incremental learning, few-shot open-set recognition, few-shot learning, zero-shot learning, and general image classification, demonstrate that our method effectively enhances the performance of hyperbolic algorithms in open-environment.
}





\maketitle

\section{Introduction}\label{sec1}
Learning in hyperbolic spaces has shown remarkable performance across various tasks~\citep{fang2023poincare,mettes2024hyperbolic,Hu_2024_CVPR,10460427}. In contrast to the Euclidean space, the hyperbolic space with a constant negative curvature is more suitable for capturing hierarchical data structures due to its exponential volume growth concerning the radius~\citep{fang2021kernel}. 
Despite their advantages, hyperbolic algorithms often rely on sufficient data. While in many real-world scenarios, data is scarce and difficult to obtain, limiting the practical performance and hindering the further development of hyperbolic algorithms~\citep{hong2023hyperbolic,hamzaoui2024hyperbolic}. 
Feature augmentation, an efficient data augmentation technique, provides a promising solution to address data scarcity by generating new features for learning in downstream tasks~\citep{Verma2018ManifoldMB,zang2021fasa,Li2020AdversarialFH}. Compared with augmentation in the raw data space (\emph{e.g.}, image or video space), augmentation in the feature space requires less memory footprint~\citep{li2021simple,zang2021fasa}, and provides more useful intra-class variations by leveraging semantic information inherent in the feature space~\citep{Wang2019ImplicitSD,Li2021TransferableSA,Li2021MetaSAugMS}. 

Most hyperbolic feature augmentation generates features in closed-environment that
only produces features for seen classes~\citep{gao2022hyperbolic}. This leads to the over-dominance issue, where features of seen classes improperly occupy most of the embedding space, conflicting with open-environment with unseen and changing numbers of classes. 
For example, class-incremental learning tasks require managing a continuous influx of new classes while preserving knowledge of previously encountered ones. Generating features exclusively for seen classes can result in allocating disproportionate embedding space to these classes, neglecting the yet-to-be-seen ones~\citep{zhu2022learning}. As new classes are introduced, they inevitably encroach upon the space allocated to old classes, causing catastrophic forgetting of previously learned information.
Similarly, in zero-shot recognition tasks, augmenting only the seen classes results in their over-dominance, undermining the task’s goal of identifying unseen categories during training. This imbalance leads to inaccurate modeling of real-world data distributions and impairs the model's ability to generalize to unseen classes.

In this paper, we propose a hyperbolic dual feature augmentation method designed for tasks of open-environment. Our method generates plausible and diverse features for both seen and unseen classes during training, preventing the embedding space from being dominated by seen classes and ensuring adequate space for incoming unseen classes.
Furthermore, introducing features of unseen classes during training helps squeeze the data distribution within each class, resulting in smaller intra-class distances and sharper classification boundaries. This, in turn, enhances the model's discriminative power. Similar findings have been observed in certain Euclidean-based methods~\citep{zhu2021class,zhu2022learning}.

Performing efficiently dual feature augmentation in the hyperbolic space involves two challenges: ensuring the plausibility and diversity of the augmented features. 
(1) Guaranteeing the plausibility of augmented features is challenging, due to the complicated and various hierarchical structures in real-world data, where the inter-class and intra-class hierarchical structures need to be preserved in feature augmentation.
(2) Increasing the diversity of augmented features is challenging, due to the data scarcity and high computational cost of complex hyperbolic operations (\emph{e.g.}, parallel transport and exponential map).
Diverse augmented features are essential, as they effectively capture underlying data distributions, but achieving this diversity can impose significant computational burdens.

In order to guarantee the plausibility of augmented features, we employ meta-learning to obtain prior knowledge about accurate distribution estimation of real-world data.
We meta-learn neural ordinary differential equations (ODEs) to estimate distributions of original seen classes and auxiliary unseen classes. 
This converts distribution estimation into a continuous optimization problem that could well capture intra-class hierarchical data structures.
Considering the fact that synthesized features for unseen classes are supposed to be of higher uncertainty than the seen classes, we introduce a hierarchy-preserving regularizer that encourages the synthesized unseen classes to be closer to the origin of the hyperbolic space and the seen classes are nearer to the boundary since the region close to the origin in the hyperbolic space corresponds to higher uncertainty.
In this way, the latent inter-class hierarchies are preserved via modeling the uncertainty, and the augmented classes are more generic and of higher plausibility. 
As for the diversity of augmented features, we derive an upper bound of the dual augmentation loss to train the hyperbolic model using an infinite number of sampled features, instead of augmenting the feature samples explicitly, thus improving efficiency. As a result, our method enhances data diversity, model generalization and efficiency without relying on complex hyperbolic operations.

We evaluate the proposed hyperbolic dual feature augmentation method on five open-environment tasks: class-incremental learning, few-shot open-set recognition, few-shot learning, zero-shot learning, and general image classification. Comprehensive experimental results demonstrate that our method improves the performance of hyperbolic algorithms when limited data is given in open-environment. 

Our main contributions are summarized as follows:
\begin{itemize}
    \item We propose a hyperbolic dual feature augmentation method (HDFA) for open-environment, which generates diverse and plausible features for both seen and unseen classes, improving hyperbolic algorithms' performance when given limited data.
    \item We introduce a regularizer that preserves the hierarchies between generated unseen classes and seen classes in open-environment.
    \item We derive an upper bound for the dual feature augmentation loss on both the seen and unseen classes. It provides a feasible computational approach to train hyperbolic models with infinite augmentation.
\end{itemize}

\section{Related Work}\label{related work}

\subsection{Augmentation Methods}  
Augmentation methods, widely used to address the challenge of data scarcity, can be categorized into raw data (\emph{e.g.}, image or video) augmentation and feature augmentation. Transformation-based (\emph{e.g.}, rotation, masking, flipping and cropping) methods are popular examples for the first category~\citep{Lim2019FastA,Gudovskiy2021AutoDORA,Gong2020KeepAugmentAS}. Another subfield of raw data augmentation is mixing-based methods. For instance, Mixup~\citep{zhang2018mixup} performs linear interpolation between two randomly sampled images, generating new samples that combine aspects of both images from a holistic perspective. Cutmix ~\citep{yun2019cutmix}, on the other hand, cuts a patch from one image and then pastes it onto the corresponding region of another image from a locality point of view. Additionally, some augmentation methods train generative models (\emph{e.g.}, Generative Adversarial Networks (GANs)~\citep{goodfellow2020generative}, Variational Autoencoders (VAEs)~\citep{kingma2013auto} and Diffusion models~\citep{ho2020denoising}) to increase data diversity by generating synthetic data for model generalization~\citep{jiang2021deceive,chadebec2022data,islam2024diffusemix}. For instance, DiffuseMix~\citep{islam2024diffusemix} leverages a diffusion model to generate samples based on tailored conditional prompts. By combining partial natural images with generated counterparts and integrating fractal patterns, DiffuseMix increases the diversity of the data while preserving the original semantics of the input image. These generating-based augmentation methods in the raw data space provide high performance gains but come with high computational complexity.

Compared with raw data augmentation, feature augmentation brings richer intra-class variants and involves less computational complexity and memory footprint. The intuition behind these works is that certain directions in the feature space correspond to meaningful semantic transformations~\citep{upchurch2017deep, bengio2013better}. Existing feature augmentation methods typically involve interpolating among feature samples~\citep{Li2020OnFN}, sampling from constructed distributions~\citep{yuan2024fakd}, and using GAN models~\citep{Lazarou2021TensorFH}. For instance, Manifold mixup~\citep{Verma2018ManifoldMB} performs interpolation among features from multiple deep network layers and propagates them further for re-sampling to enrich feature representations. ISDA~\citep{Wang2019ImplicitSD} performs feature level augmentation by randomly sampling feature vectors from a zero-mean normal distribution obtained with the help of a pre-computed category-specific covariance matrix of extracted features. AFHN~\citep{Li2020AdversarialFH} uses conditional Wasserstein Generative Adversarial networks~\citep{arjovsky2017wasserstein} to generate diverse and discriminative features conditioned on the few labeled samples.
Unlike the above methods that focus on augmentation in the Euclidean space, our approach explores feature augmentation in the hyperbolic space, which better captures the hierarchical structures inherent in real-world data.


\subsection{Hyperbolic Learning}
Hyperbolic space, regarded as the continuous analogue of discrete trees, is particularly suited for modeling data with hierarchical structures and power-law distributions.
This property enables richer representations and has driven the successful development of learning algorithms in hyperbolic spaces for various applications, particularly benefiting out-of-distribution generalization~\citep{guo2022clipped, hong2023curved}, uncertainty quantification~\citep{chen2023hyperbolic}, and tasks in low-data regimes, such as few-shot learning~\citep{Anvekar_2023_CVPR, hamzaoui2024hyperbolic} and zero-shot recognition~\citep{xu2022meta, hong2023hyperbolic}.
Existing efforts of exploiting hyperbolic spaces mainly focus on the model level (\emph{i.e.}, designing/improving hyperbolic models) and the data level. Current works of the first category focus on extending traditional neural networks to hyperbolic spaces, such as hyperbolic convolutional
networks~\citep{bdeir2024fully}, hyperbolic residual networks~\citep{van2023poincare}, hyperbolic graph networks~\citep{xue2024residual} and hyperbolic Transformer networks~\citep{yang2024hypformer}. 

The second category, to which our method belongs, focuses on acquiring
more representative and generalized features via data augmentation techniques tailored for hyperbolic spaces~\citep{sawhney2021hypmix, gao2022hyperbolic}.
For instance, Sawhney et al.~\citep{sawhney2021hypmix} introduced HypMix as interpolative operations in hyperbolic spaces to generate novel samples, better capturing the complex geometry of latent representations. Gao et al.~\citep{gao2022hyperbolic} proposed a hyperbolic feature augmentation method that overcomes overfitting by generating class-identity-preserving features in the hyperbolic space. One common property of the above data augmentation methods in hyperbolic spaces is that they generate synthetic samples in a way that is consistent with the data distribution of the seen classes. These methods neglect the existence of unseen classes outside the original distribution, resulting in an erroneous approximation of the real data distribution and leading to the over-dominance of seen classes in open-environment.
Different from these methods, our method generates
diverse and plausible features for both seen and
unseen classes, addressing over-dominance in the
embedding space and enhancing the generalization of hyperbolic algorithms in
open-environment.

\section{Preliminaries}\label{preliminaries}
As a counterpart in the Euclidean space, a hyperbolic space is a smooth Riemannian manifold with a constant negative curvature~\citep{lee2006riemannian}. Hyperbolic space has five isometric models, and this paper operates on the most commonly used model of hyperbolic geometry in deep learning, namely the Poincar\'e ball model~\citep{cannon1997hyperbolic}. A $d$-dimensional Poincar\'e ball model with constant negative curvature $c$ is defined as $\mathcal{M}^{d, c}=\left\{\boldsymbol{x} \in \mathbb{R}^d,-c\|\boldsymbol{x}\|<1\right\}$, where $\|\cdot\|$ is the Euclidean norm. For $\boldsymbol{x} \in \mathcal{M}^{d, c}$, $T_{\boldsymbol{x}
}\mathcal{M}^{d, c}$ denotes its tangent space that contains all tangent vectors to $\mathcal{M}^{d, c}$ at $\boldsymbol{x}$. In the following, we will introduce the used hyperbolic operations.

\textbf{Addition.} For vectors $\boldsymbol{x},\boldsymbol{y} \in \mathcal{M}^{d, c}$, their addition is 
\begin{equation}\small
    \label{addition}
    \begin{aligned}
        \boldsymbol{x} \oplus_c \boldsymbol{y}=\frac{\left(1-2 c\langle\boldsymbol{x}, \boldsymbol{y}\rangle_2-c\|\boldsymbol{y}\|^2\right) \boldsymbol{x}+\left(1+c\|\boldsymbol{x}\|^2\right) \boldsymbol{y}}{1-2 c\langle\boldsymbol{x}, \boldsymbol{y}\rangle_2+c^2\|\boldsymbol{x}\|^2\|\boldsymbol{y}\|^2} .
    \end{aligned}
\end{equation}

\begin{figure*}[t]
\centering
\includegraphics[width=1.95\columnwidth]{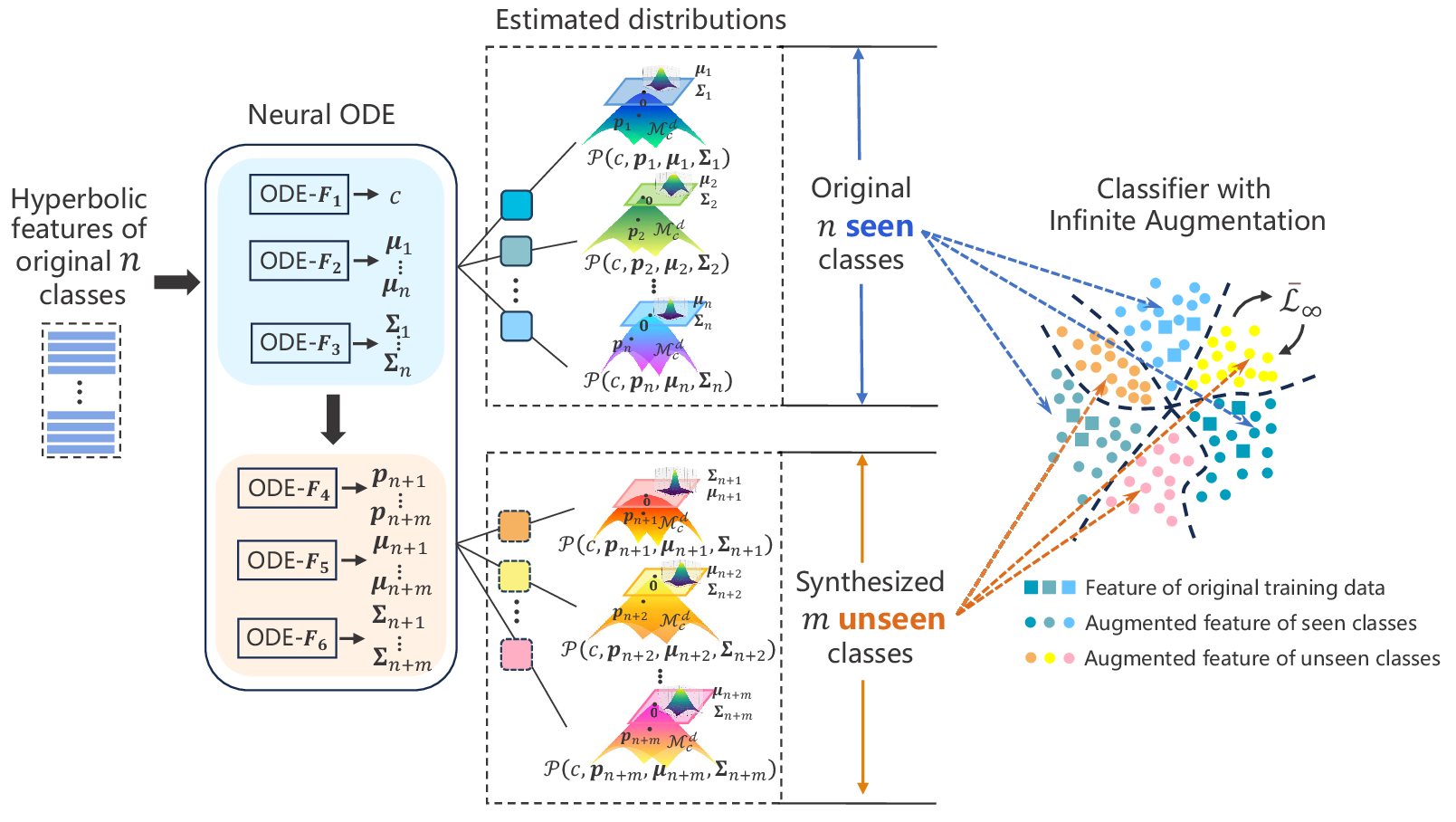}
\caption{A conceptual diagram of dual feature augmentation in the hyperbolic space.}
\label{dualaug}


\end{figure*}


\textbf{Distance measure.} For two points $\boldsymbol{x},\boldsymbol{y} \in \mathcal{M}^{d, c}$, their distance is 
\begin{equation}\small
    \begin{aligned}
        d^c(\boldsymbol{x}, \boldsymbol{y})=\frac{1}{\sqrt{|c|}} \cosh ^{-1}\left(1-2 c \frac{\|\boldsymbol{x}-\boldsymbol{y}\|^2}{\left(1+c\|\boldsymbol{x}\|^2\right)\left(1+c\|\boldsymbol{y}\|^2\right)}\right) .
    \end{aligned}   
    \label{dis}
\end{equation}

\textbf{Exponential and logarithmic map.} For points $\boldsymbol{x},\boldsymbol{y} \in \mathcal{M}^{d, c}$, tangent vector $\boldsymbol{u} \in T_{\boldsymbol{x}
}\mathcal{M}^{d, c}$, the exponential map $\operatorname{expm}_{\boldsymbol{x}}^c : T_{\boldsymbol{x}
}\mathcal{M}^{d, c} \rightarrow \mathcal{M}^{d, c}$ and the logarithmic map $\operatorname{logm}_{\boldsymbol{x}}^c: \mathcal{M}^{d, c}\rightarrow T_{\boldsymbol{x}
}\mathcal{M}^{d, c}$
are defined as
\begin{equation}\small
    \begin{aligned} \operatorname{expm}_{\boldsymbol{x}}^c(\boldsymbol{u})=\boldsymbol{x} \oplus_c\left(\tanh \left(\sqrt{|c|} \frac{\lambda_{\boldsymbol{x}}^c\|\boldsymbol{u}\|}{2}\right) \frac{\boldsymbol{u}}{\sqrt{|c|} \| \boldsymbol{u} \mid}\right),
    \end{aligned}   
\end{equation}

\begin{align}
    \begin{aligned} \operatorname{logm}_{\boldsymbol{x}}^c(\boldsymbol{y})=\frac{2}{\sqrt{|c|} \lambda_{\boldsymbol{x}}^c} \tanh ^{-1}&\left(\sqrt{|c|}\left\|-\boldsymbol{x} \oplus_c \boldsymbol{y}\right\|\right) \\
    &\times \frac{-\boldsymbol{x} \oplus_c \boldsymbol{y}}{\left\|-\boldsymbol{x} \oplus_c \boldsymbol{y}\right\|},&
    \end{aligned}
\end{align}
where $\lambda_{\boldsymbol{x}}^c=2 /\left(1+c\|\boldsymbol{x}\|^2\right)$ is the conformal factor. The exponential map $\operatorname{expm}_{\boldsymbol{x}}^c(\boldsymbol{u}$ projects vector $\boldsymbol{u}$ from the tangent space $T_{\boldsymbol{x}
}\mathcal{M}^{d, c}$ to the manifold $\mathcal{M}^{d, c}$, and the logarithmic map $\operatorname{logm}_{\boldsymbol{x}}^c(\boldsymbol{y})$ maps vector $\boldsymbol{y}$ from the manifold to the tangent space. The exponential map in the tangent space $T_{\boldsymbol{0}
}\mathcal{M}^{d, c}$ at the origin is simplified to $\operatorname{expm}_{\boldsymbol{0}}^c(\boldsymbol{u})=\tanh (\sqrt{|c|}\|\boldsymbol{u}\|) \frac{\boldsymbol{u}}{\sqrt{|c|}|\boldsymbol{u}| \mid}$, and the logarithmic map $\operatorname{logm}_{\boldsymbol{0}}^c$ in $T_{\boldsymbol{0}
}\mathcal{M}^{d, c}$ is simplified to $\operatorname{logm}_{\boldsymbol{0}}^c(\boldsymbol{y})=\tanh ^{-1}(\sqrt{|c|}\|\boldsymbol{y}\|) \frac{\boldsymbol{y}}{\sqrt{|c|} \| \boldsymbol{y} \mid}$.

\textbf{Wrapped Normal Distribution.} 
The wrapped normal distribution~\citep{mathieu2019continuous,li2022curvature} has been developed to model data in hyperbolic spaces, showing effectiveness in multiple domains.
Its density function is defined as
\begin{equation}\small
    \begin{aligned}
        \mathcal{P}(\boldsymbol{x} \mid c, \boldsymbol{p}, \boldsymbol{\mu}, \boldsymbol{\Sigma})=\mathcal{N}&\left(\lambda_{\boldsymbol{p}}^c \operatorname{logm} _{\boldsymbol{p}}^c(\boldsymbol{x}) \mid \boldsymbol{\mu}, \boldsymbol{\Sigma}\right)\\
        & \times \left(\frac{\sqrt{|c|} d^c(\boldsymbol{p}, \boldsymbol{x})}{\sinh \left(\sqrt{|c|} d^c(\boldsymbol{p}, \boldsymbol{x})\right)}\right).
    \end{aligned}
\end{equation}
Here, $c$ denotes the curvature of the underlying hyperbolic space, $\boldsymbol{\mu}$ and $\boldsymbol{\Sigma} \in \mathbb{R}^{d \times d} $ denote the mean and covariance matrix of features, and $\boldsymbol{p} \in \mathcal{M}^{d, c}$ denotes the prototype of data. 

Generating a hyperbolic feature $\boldsymbol{x}$ from a wrapped normal distribution $\mathcal{P}(c, \boldsymbol{p}, \boldsymbol{\mu}, \boldsymbol{\Sigma})$ includes three steps. (1) Sample a vector $\widehat{\boldsymbol{v}}$ from the normal distribution $\mathcal{N}(\boldsymbol{\mu}, \boldsymbol{\Sigma})$ in the tangent space $T_{\mathbf{0}} \mathcal{M}^{d, c}$ at the origin. (2) Transport $\widehat{\boldsymbol{v}}$ from $T_0 \mathcal{M}^{d, c}$ to $\boldsymbol{v}$ in the tangent space $T_p \mathcal{M}^{d, c}$ at the prototype $\boldsymbol{p}$, $\boldsymbol{v}=\mathrm{PT}_{\mathbf{0} \rightarrow \boldsymbol{p}}^c(\widehat{\boldsymbol{v}})$. (3) Map $\boldsymbol{v}$ to the manifold $\mathcal{M}^{d, c}$ to obtain $\boldsymbol{x}$, using $\boldsymbol{x} = \operatorname{expm}_{\boldsymbol{p}}^c(\boldsymbol{v})$. Since sampling $\widehat{\boldsymbol{v}}$ from $\mathcal{N}(\boldsymbol{\mu}, \boldsymbol{\Sigma})$ in step (1) is not a differentiable process, gradients cannot be backpropagated to $\boldsymbol{\mu}$ and $\boldsymbol{\Sigma}$. To address this, we use a reparameterization trick. We sample a vector $\boldsymbol{\epsilon}$ from the standard normal distribution $\mathcal{N}(\mathbf{0}, \boldsymbol{I})$ and transform it into $\widehat{\boldsymbol{v}}$ via $\widehat{\boldsymbol{v}} = \boldsymbol{\mu} + \boldsymbol{L} \boldsymbol{\epsilon}$, where $\boldsymbol{L} \boldsymbol{L}^{\top} = \boldsymbol{\Sigma}$. 

\textbf{Neural Ordinary Differential Equation.}
Neural ordinary differential equation (neural ODE), introduced by~\citep{chen2018neural}, is a kind of
continuous-time model, offering an elegant approach to model the optimization of hidden states as a continuous process.
It governs the optimization process by solving the solution of an ordinary differential equation in deep learning. 


Concretely, a hidden state $h$ is optimized via an integral term, $h^T = h^0+\int_{t=0}^T \nabla {h}^t d t $, and $\nabla {h}^t$ denotes the gradient flow of $h$ at time $t$. $\nabla {h}^t$ is usually computed via a neural network $F$, $\nabla {h}^t = \frac{d h^t}{d t}=F(h^t, t)$.
Neural ODEs leverage an ODE solver (such as the Runge-Kutta method~\citep{doi:10.1137/1032091}), denoted by $\operatorname{ODESolve}(\cdot)$, to solve the integral term, ${h}^T=\operatorname{ODESolve}\left({h}^0, F, T\right)$.
It ensures a memory-efficient backpropagation, maintaining a constant $O(1)$ memory cost.

Neural ODEs allow for the introduction of time as a continuous control variable, enabling smoother and more flexible optimization processes compared to traditional discrete optimization methods.
Neural ODEs have been successfully applied across various domains, including time-series prediction~\citep{kidger2020neural}, knowledge graph forecasting~\citep{han2021learning}, and few-shot learning~\citep{zhang2022metanode}. Owing to their ability to dynamically model the continuous evolution of states with fine granularity, neural ODEs provide an effective and flexible approach to capture the complex structures inherent in real-world data.



\section{Method}
\subsection{Formulation}
In this paper, we propose the hyperbolic dual feature augmentation method for open-environment. It generates diverse and plausible features for both seen and unseen classes in the hyperbolic space. 
Firstly, we meta-learn neural ODEs to estimate the wrapped normal distributions for seen classes and unseen classes, where distributions of seen classes are firstly estimated and combined for unseen classes.
Next, we introduce a regularizer to preserve the hierarchical relationships between the synthesized unseen classes and seen classes according to the uncertainty information of classes. 
Finally, we derive an upper bound $\overline{\mathcal{L}}_{\infty}$ of the loss function with infinite augmentation of seen and unseen classes to train the classifier. Figure~\ref{dualaug} shows the conceptual diagram of our hyperbolic dual feature augmentation.

\subsection{Distribution Estimation for Seen Classes}\label{sec:ODE}
We leverage neural ODEs to estimate the wrapped normal distributions of seen classes. Specifically, we compute their prototypes by averaging the data, and estimate the curvature, mean, and covariance matrices by using neural ODEs. All classes have a common estimated curvature and individual mean and covariance matrices.
The estimation of the common curvature $c$, mean $\boldsymbol{\mu}_{i}$ for the $i$-th seen class and individual matrix $\boldsymbol{L}_{i}$ for the $i$-th seen class can be cast as iterative optimization processes, 
\begin{equation}
\begin{aligned}
    \left\{\begin{array}{l}
    c^{t+1}=c^t+\nabla c^t \\
\boldsymbol{\mu}_{i}^{t+1}=\boldsymbol{\mu}_{i}^t+\nabla \boldsymbol{\mu}_{i}^t \\
\boldsymbol{L}_{i}^{t+1}=\boldsymbol{L}_{i}^t+\nabla \boldsymbol{L}_{i}^t
\end{array}\right. ,
\end{aligned}
\label{eq:gradient}
\end{equation}
where $\nabla c^t$, $\nabla \boldsymbol{\mu}_{i}^t$, $\nabla \boldsymbol{L}_{i}^t$ denote the gradient flows of $c$, $\boldsymbol{\mu}_{i}$ and $\boldsymbol{L}_{i}$ at time $t$, respectively. We view the above iterative optimization processes as continues-time processes of ODEs, and use three neural networks $F_{1\sim3}$ to estimate the gradient flows at time $t$, denoted by 
\begin{equation}
\begin{aligned}
    \left\{\begin{array}{l}
 \nabla c^t=\frac{d c^t}{d t}=F_{1}\left(c^t, t\right)\\
 \nabla \boldsymbol{\mu}_{i}^t=\frac{d \boldsymbol{\mu}_{i}^t}{d t}=F_{2}\left(\boldsymbol{\mu}_{i}^t, t\right)\\
 \nabla \boldsymbol{L}_{i}^t=\frac{d \boldsymbol{L}_{i}^t}{d t}=F_{3}\left(\boldsymbol{L}_{i}^t, t\right)
\end{array}\right. .
\end{aligned}
\label{eq:gradient2}
\end{equation}
Thus, given gradient flow networks $F_{1\sim3}$ and the initial values $c^0$, $\boldsymbol{\mu}_{i}^0$ and $\boldsymbol{L}_{i}^0$, the distribution parameters can be obtained by solving the neural ODEs at the last step $T$ with an integral term,
\begin{equation}\small
\begin{aligned}
    \left\{\begin{array}{l}
    {c}^T={c}^0+\int_{t=0}^T F_1\left({c}^t, t\right) d t \\
    {\boldsymbol{\mu}_i}^T={\boldsymbol{\mu}_i}^0+\int_{t=0}^T F_2\left({\boldsymbol{\mu}_i}^t, t\right) d t
    \\{\boldsymbol{L}_i}^T={\boldsymbol{L}_i}^0+\int_{t=0}^T F_3\left({\boldsymbol{L}_i}^t, t\right) d t
\end{array}\right. .
\end{aligned}
\label{eq:ode2}
\end{equation}
Here, $\boldsymbol{\mu}_i^0$ and $\boldsymbol{L}_i^0$ are obtained by computing the mean of given data, and ${c}^0$ is set as a fixed value. Similar to work~\citep{gao2022hyperbolic}, we utilize the Runge-Kutta method~\citep{doi:10.1137/1032091} denoted by $\operatorname{ODESolve}(\cdot)$ to solve the integral term in Eq.~\eqref{eq:ode2}, 
\begin{equation}\small
    \begin{aligned}
        \left\{\begin{array}{l}
        {c}^T=\operatorname{ODESolve}\left({c}^0, F_{1}, T\right)\\{\boldsymbol{\mu}_i}^T=\operatorname{ODESolve}\left({\boldsymbol{\mu}_i}^0, F_{2}, T\right)\\{\boldsymbol{L}_i}^T=\operatorname{ODESolve}\left({\boldsymbol{L}_i}^0, F_{3}, T\right)
        \end{array}\right. .
    \end{aligned}
\label{eq:ode}
\end{equation}

\begin{figure*}[t]
    \centering
    \includegraphics[width=1.6\columnwidth]{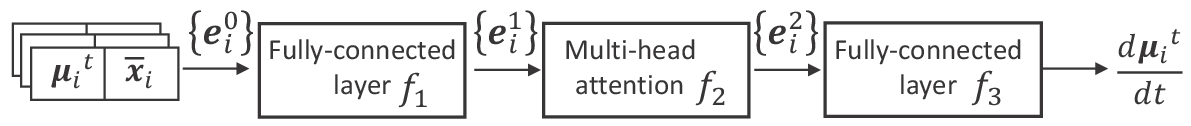}
    \caption{Architecture of the gradient flow network.}
    \label{fig:GFN}
\end{figure*}

The architecture of the gradient flow network used for estimating the distribution parameters is shown in Figure~\ref{fig:GFN}. 
Taking the estimation of parameter $\boldsymbol{\mu}_{i}$ as an example, 
the network takes 
$\boldsymbol{\mu}_{i}^{t}$ at the current time and 
features belonging to the $i$-th class as inputs, and generates the gradient $\frac{d \boldsymbol{\mu}_{i}^t}{d t}$ via three layers. Concretely, the distribution parameter $\boldsymbol{\mu}_{i}^t$ and the mean $\overline{\boldsymbol{x}}_i$ of features belonging to the $i$-th seen class are concatenated as an estimation representation $\boldsymbol{e}_i^{0}=\left[\boldsymbol{\mu}_{i}^t, \overline{\boldsymbol{x}}_i\right]$. 
We first use a fully-connected layer $f_{1}$ to embed $\boldsymbol{e}_i^{0}$ as
\begin{equation}\small
    \begin{aligned}
        \boldsymbol{e}_i^{1}=f_{1}\left(\boldsymbol{e}_i^{0}\right).
    \end{aligned}
\end{equation}
Then, a self-attention layer $f_{2}$ is used to perform interaction among different classes,
\begin{equation}\small
    \begin{aligned}
       \boldsymbol{e}_i^{2}=f_{2}\left(\left\{\boldsymbol{e}_i^{1}\right\}_{i=1}^n\right),
    \end{aligned}
\end{equation}
where the query, key, and value of the self-attention layer $f_{2}$ are all $\left\{\boldsymbol{e}_i^{1}\right\}_{i=1}^n$ of the $n$ seen classes in total. Finally, a fully connected layer $f_{3}$ is utilized to generate the gradient,
\begin{equation}\small
    \begin{aligned}
       \frac{d \boldsymbol{\mu}_{i}^t}{d t}=f_{3}\left(\left\{\boldsymbol{e}_i^{2}\right\}\right) .
    \end{aligned}
\end{equation}

\subsection{Synthesize Unseen Classes by Perturbation}
\label{sec:novel class}
\begin{figure}[t]
\begin{subfigure}[b]{0.45\columnwidth}
        \centering
        \includegraphics[width=1.1\columnwidth]{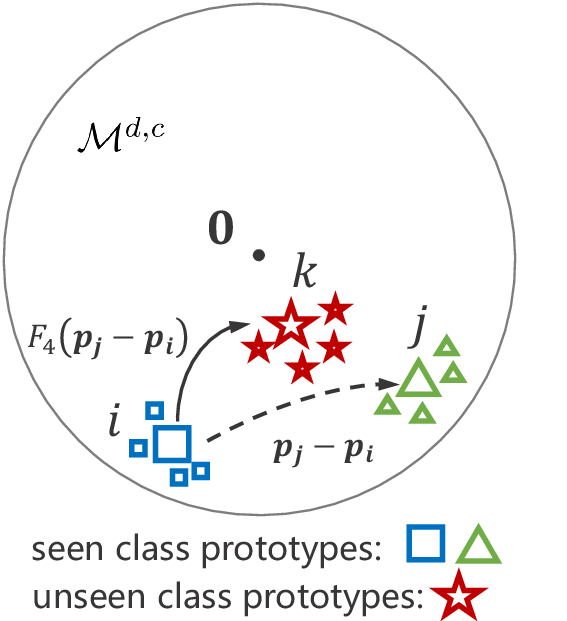}
        \caption{Estimate the prototype of unseen class $k$ by perturbation.}
        \label{fig:classaug}
    \end{subfigure}
    \begin{subfigure}[b]{0.45\columnwidth}
        \centering
        \includegraphics[width=1.1\columnwidth]{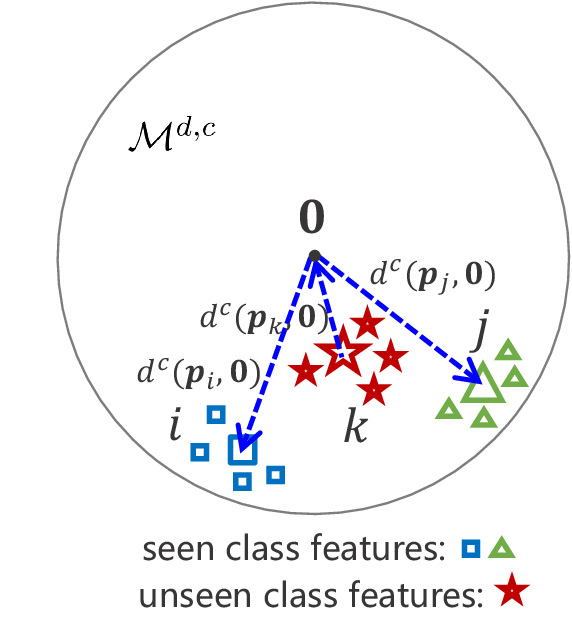}
        \caption{$R_{Hier}$: features of class $k$ should be closer to the center.}
        \label{fig:classaug2}
    \end{subfigure}
\caption{Unseen class augmentation.} 
\end{figure}

We estimate perturbation directions that are applied to distributions of seen classes to synthesize unseen classes.
Given two seen classes $i$ and $j$ with their distributions $\mathcal{P}\left(c, \boldsymbol{p}_i, \boldsymbol{\mu}_i, \boldsymbol{\Sigma}_i\right)$ and $\mathcal{P}\left(c, \boldsymbol{p}_j, \boldsymbol{\mu}_j, \boldsymbol{\Sigma}_j\right)$, we estimate the prototype, mean, and covariance of unseen classes by using neural ODEs, which could be represented by a uniform denotation $\delta$. 
We add the perturbation direction $\delta$ to the class $i$ to synthesize the unseen class $k$: $\mathcal{P}\left(c, \boldsymbol{p}_k, \boldsymbol{\mu}_k, \boldsymbol{\Sigma}_k\right)$, as shown in Figure~\ref{fig:classaug}.

Concretely, we first compute the differences between $\mathcal{P}\left(c, \boldsymbol{p}_i, \boldsymbol{\mu}_i, \boldsymbol{\Sigma}_i\right)$ and $\mathcal{P}\left(c, \boldsymbol{p}_j, \boldsymbol{\mu}_j, \boldsymbol{\Sigma}_j\right)$ as the initial perturbation $\delta^0$, including $\delta_{\boldsymbol{p}}^0=\boldsymbol{p}_j-\boldsymbol{p}_i$, $\delta_{\boldsymbol{\mu}}^0=\boldsymbol{\mu}_j-\boldsymbol{\mu}_i$ and $\delta_{\boldsymbol{L}}^0=\boldsymbol{L}_j-\boldsymbol{L}_i$. Then, we use neural ODEs to estimate perturbation directions for prototype, mean, and covariance matrices via $T$ steps, 
\begin{equation}
    \begin{aligned}
        \left\{\begin{array}{l}\delta_{\boldsymbol{p}}^T=\operatorname{ODESolve}\left(\delta_{\boldsymbol{p}}^0, F_4, T\right)\\
\delta_{\boldsymbol{\mu}}^T=\operatorname{ODESolve}\left(\delta_{\boldsymbol{\mu}}^0, F_5, T\right)\\ \delta_{\boldsymbol{L}}^T=\operatorname{ODESolve}\left(\delta_{\boldsymbol{L}}^0, F_6, T\right)
        \end{array}\right. ,
    \end{aligned}
    \label{eq:per}
\end{equation}
where three gradient flow networks $F_4$, $F_5$, and $F_6$ are used to produce gradient flows for the prototype, mean, and covariance matrices, denoted by
\begin{equation}
    \begin{aligned}
        \left\{\begin{array}{l}
\frac{d \delta_{\boldsymbol{p}}^t}{d t}=F_4\left(\delta_{\boldsymbol{p}}^t, t\right)\\
\frac{d \delta_{\boldsymbol{\mu}}^t}{d t}=F_5\left(\delta_{\boldsymbol{\mu}}^t, t\right)\\ \frac{d \delta_{\boldsymbol{L}}^t}{d t}=F_6\left(\delta_{\boldsymbol{L}}^t, t\right)
        \end{array}\right. .
    \end{aligned}
    \label{eq:per2}
\end{equation}
Thus, the distribution of the unseen class $k$ is computed as
\begin{equation}
    \begin{aligned}
    \left\{\begin{array}{l}
    \boldsymbol{p}_k = \boldsymbol{p}_i +\delta_{\boldsymbol{p}}^T \\ \boldsymbol{\mu}_k=\boldsymbol{\mu}_i+\delta_{\boldsymbol{\mu}}^T \\\boldsymbol{L}_k=\boldsymbol{L}_i+\delta_{\boldsymbol{L}}^T
    \end{array}\right. .
 \end{aligned}
 \label{eq:unseen_estimation}
\end{equation}


Assuming the original training dataset $\mathcal{D}_t$ contains $n$ seen classes, we can synthesize $m=n(n-1)/2$ unseen classes using the above method. Consequently, the original $n$-class 
problem in the current task is extended to a $(n + m)$-class problem. By exposing the model to a greater variety of classes beyond the original distribution during training, our method reserves room for potential incoming classes in the embedding space, avoiding the over-dominance of seen classes. Additionally, this encourages the model to learn sharp classification boundaries, enhancing its discriminative power.

\subsection{Hierarchy-Preserving Regularizer}
We introduce a hierarchy-preserving regularizer $R_{Hier}$ to guarantee the plausibility of augmented features. 
Considering that the unseen classes are supposed to be of higher uncertainty than the seen classes, $R_{Hier}$ preserves the latent hierarchies between unseen classes and seen class pairs by leveraging hyperbolic uncertainty information~\citep{chen2023hyperbolic}, ensuring that features with higher uncertainty are positioned closer to the center of the Poincar\'e ball. It is formally defined as 
\begin{equation}\small
    \begin{aligned}
        R_{Hier} &= 2*d^c(\boldsymbol{p}_k, \boldsymbol{0})-d^c(\boldsymbol{p}_i, \boldsymbol{0})-d^c(\boldsymbol{p}_j, \boldsymbol{0})+\\
        &{\rm max}\Big(0,\gamma \big(d^c(\boldsymbol{p}_i, \boldsymbol{0})+d^c(\boldsymbol{p}_j, \boldsymbol{0})\big)-d^c(\boldsymbol{p}_k, \boldsymbol{0})\Big).
    \end{aligned}
\end{equation}
Here, $d^c(\boldsymbol{p}_k, \boldsymbol{0})$ denotes the hyperbolic distance from the embedding of $\boldsymbol{p}_k$ to the origin $\boldsymbol{0}$, which is used to measure the hyperbolic uncertainty~\citep{atigh2022hyperbolic,chen2023hyperbolic}.
According to Eq.~\eqref{dis}, $d^c(\boldsymbol{p}_k, \boldsymbol{0})$ is computed by
\begin{equation}\small
    \begin{aligned}
        d^c(\boldsymbol{p}_k, \boldsymbol{0})=\frac{1}{\sqrt{|c|}} \cosh ^{-1}\left(1-2 c \frac{\|\boldsymbol{p}_k\|^2}{\left(1+c\|\boldsymbol{p}_k\|^2\right)}\right).
    \end{aligned}
\end{equation}
It serves as a measure of how confidently or certainly the point (\emph{i.e.}, feature of a certain class) will be classified. When
points are classified with high confidence, their embeddings
should be far away from the center of the Poincar\'e ball. By contrast, for points
classified with high uncertainty, their embeddings are closer
to the center. For example, in Figure~\ref{fig:classaug2}, $R_{Hier}$ forces the hyperbolic
distance of the synthesized unseen class $k$ to be shorter than that of the seen classes $i$ and $j$, indicating that class $k$ are more generic. Meanwhile, $R_{Hier}$ employs the second term ${\rm max}\Big(0,\gamma \big(d^c(\boldsymbol{p}_i, \boldsymbol{0})+d^c(\boldsymbol{p}_j, \boldsymbol{0})\big)-d^c(\boldsymbol{p}_k, \boldsymbol{0})\Big)$ to prevent $\boldsymbol{p}_k$ from getting too close to the origin $\boldsymbol{0}$. $\gamma$ is a hyperparameter used to control the level of the uncertainty of $\boldsymbol{p}_k$, and we choose its value from the interval [0.01, 0.25] in this work.

\begin{algorithm} [t]
    \caption{Training process of HDFA.}
    \label{alg:train} 
    \textbf{Input}: The base dataset $\mathcal{D}$.\\
    \textbf{Output}: The learned gradient flow networks $F_1$, $F_2$, $F_3$, $F_4$, $F_5$ and $F_6$.
    \begin{algorithmic}[1] 
        \WHILE{Not converged}
        \STATE
        Randomly select a few data from $\mathcal{D}$ as training data $\mathcal{D}_t$, and the rest as validation data $\mathcal{D}_v$.
        \STATE
        Use gradient flow networks $F_{1\sim 6}$ to estimate the distributions of seen classes in $\mathcal{D}_t$ and the unseen classes via Eq.~\eqref{eq:ode} and Eq.~\eqref{eq:unseen_estimation}.
        \WHILE{Not converged}
        \STATE
        Use the estimated distributions to update the classifier $\boldsymbol{W}$ by minimizing the loss function in Eq.~\eqref{loss:inifite11}.
        \ENDWHILE
        \STATE
        Update $F_{1\sim 6}$ by minimizing the loss of the updated classifier on $\mathcal{D}_v$ in Eq.~\eqref{loss:outer}.   
        \ENDWHILE
  \end{algorithmic}
\end{algorithm}

\subsection{Learning with Infinite Data Augmentation}
We utilize a distance-based classifier $\boldsymbol{W}=\left[\boldsymbol{w}_1, \cdots, \boldsymbol{w}_{n+m}\right]$ for features of seen and unseen classes in the hyperbolic space, where $\boldsymbol{w}_j \in \mathcal{M}^{d, c}$ denotes the weight of the $j$-th class.
A hyperbolic feature $\boldsymbol{x}$ is classified by assigning it to the closest weight of classes, and the probability that $\boldsymbol{x}$ belongs to the $j$-th class is 
\begin{equation}\small
    \begin{aligned}
        p(\widehat{y}=j \mid \boldsymbol{x})=\frac{\exp \left(-d^c\left(\boldsymbol{x}, \boldsymbol{w}_j\right)\right)}{\sum_{j^{\prime}=1}^{n+m} \exp \left(-d^c\left(\boldsymbol{x}, \boldsymbol{w}_{j^{\prime}}\right)\right)},
    \end{aligned}
\end{equation}
where $\widehat{y}$ is the prediction of the classifier.

We sample features of seen and unseen classes from the estimated wrapped distributions to jointly train the classifier. For the $j$-th class, we generate $h$ augmented features $\left\{\boldsymbol{s}_{j, i}\right\}_{i=1}^{h}$ from its
distribution, \emph{i.e.}, $\boldsymbol{s}_{j, i} \sim \mathcal{P}\left(c, \boldsymbol{p}_j, \boldsymbol{\mu}_j, \boldsymbol{\Sigma}_j\right)$. Then, features sampled from $n$ seen and $m$ unseen classes are together fed to the classifier for minimizing the following cross-entropy loss function,
\begin{equation}\footnotesize
    \begin{aligned}
        \mathcal{L}(\boldsymbol{W})=\frac{1}{n+m} \sum_{j=1}^{n+m} \frac{1}{h} \sum_{i=1}^h-\log \frac{\exp \left(-d^c\left(\boldsymbol{s}_{j, i}, \boldsymbol{w}_j\right)\right)}{\sum_{j^{\prime}=1}^{n+m} \exp \left(-d^c\left(\boldsymbol{s}_{j, i}, \boldsymbol{w}_{j^{\prime}}\right)\right)} .
    \end{aligned}
    \label{LW}
\end{equation}

Although a large value of $h$ brings more diversity and helps the model learn class invariance, the computational cost of complex hyperbolic operations (\emph{e.g.}, parallel transport and exponential map) makes computing Eq.~\eqref{LW} inefficient. Instead, we implicitly generate infinite instances in the hyperbolic feature space to train the classifier efficiently. As $h$ approaches infinity, the loss function is defined by the expectation of the cross-entropy,
\begin{equation}\footnotesize
    \begin{aligned}
        \mathcal{L}_{\infty}(\boldsymbol{W})=\frac{1}{n+m} \sum_{j=1}^{n+m} \mathbb{E}_i\left[-\log \frac{\exp \left(-d^c\left(\boldsymbol{s}_{j, i}, \boldsymbol{w}_j\right)\right)}{\sum_{j^{\prime}=1}^{n+m} \exp \left(-d^c\left(\boldsymbol{s}_{j, i}, \boldsymbol{w}_{j^{\prime}}\right)\right)}\right].
    \end{aligned}
    \label{LW_infinite}
\end{equation}
Then, we derive an upper bound $ \overline{\mathcal{L}}_{\infty}(\boldsymbol{W})$ of $ \mathcal{L}_{\infty}(\boldsymbol{W})$, as shown in Proposition~\ref{proposit}. In this case, we train the classifier by minimizing the upper
bound $ \overline{\mathcal{L}}_{\infty}(\boldsymbol{W})$, rather than the original loss function $ \mathcal{L}_{\infty}(\boldsymbol{W})$.

\begin{proposition}
\label{proposit}
    Given infinite data $\boldsymbol{s}_{j, i} \sim \mathcal{P}\left(c, \boldsymbol{p}_j, \boldsymbol{\mu}_j, \boldsymbol{\Sigma}_j\right)$, an upper bound of the loss function $ \mathcal{L}_{\infty}$ in Eq.~\eqref{LW_infinite} is computed by
    \begin{equation}\small
        \begin{aligned}
            &\overline{\mathcal{L}}_{\infty}(\boldsymbol{W})=\frac{1}{{n+m}} \sum_{j=1}^{n+m}-\log \frac{\exp\Big(\boldsymbol{w}_j^{\top}(\widehat{\boldsymbol{p}}_j+\boldsymbol{\mu}_i)\Big)}{\xi}, 
        \end{aligned}
        \label{loss:inifite11}
    \end{equation}            
where $\xi=\sum_{j^{\prime}=1}^{n+m}\exp \Big(\boldsymbol{w}_{j^{\prime}}^{\top}\left(\widehat{\boldsymbol{p}}_j+\boldsymbol{\mu}_i\right)+(\boldsymbol{w}_{j^{\prime}}^{\top} \boldsymbol{w}_{j^{\prime}}-\boldsymbol{w}_j^{\top} \boldsymbol{w}_j)$$+ \frac{1}{2}\left(\boldsymbol{w}_{j^{\prime}}-\boldsymbol{w}_j\right)^{\top} \boldsymbol{\Sigma}_j\left(\boldsymbol{w}_{j^{\prime}}-\boldsymbol{w}_j\right)\Big)$. $\widehat{\boldsymbol{p}}_j$ denotes the corresponding point of $\boldsymbol{p}_j$ on the tangent space, that is, $\boldsymbol{p}_j=\operatorname{expm} _{\boldsymbol{0}}^c\left(\widehat{\boldsymbol{p}}_j\right)$, and $\widehat{\boldsymbol{p}}_j$ can be directly computed in the tangent space at the origin.
\end{proposition}

\noindent \textit{Proof}. We rewrite the loss function of the infinite hyperbolic dual feature augmentation in Eq.~\eqref{LW_infinite} as
\begin{align}
&\mathcal{L}_{\infty}(\boldsymbol{W})\notag =\frac{1}{{n+m}} \sum_{j=1}^{n+m} \mathbb{E}_i\bigg[ \\
& -\log \Big(\frac{\exp \big(-d^c(\boldsymbol{s}_{j, i}, \boldsymbol{w}_j)\big)}{\sum_{j^{\prime}=1}^{n+m} \exp \big(-d^c(\boldsymbol{s}_{j, i}, \boldsymbol{w}_{j^{\prime}})\big)}\Big)\bigg] \notag\\
& =\frac{1}{{n+m}} \sum_{j=1}^{n+m} \mathbb{E}_i\bigg[\log \Big(\sum_{j^{\prime}=1}^{n+m} \exp \big(d^c(\boldsymbol{s}_{j, i}, \boldsymbol{w}_j)\notag\\
& \hspace{13em} -d^c(\boldsymbol{s}_{j, i}, \boldsymbol{w}_{j^{\prime}})\big)\Big)\Bigg] \notag\\
& \leq \frac{1}{{n+m}} \sum_{j=1}^{n+m} \log \Bigg(\sum_{j^{\prime}=1}^{n+m} \mathbb{E}_i\Big[\exp \big(d^c(\boldsymbol{s}_{j, i}, \boldsymbol{w}_j)\notag\\
& \hspace{10em} -d^c(\boldsymbol{s}_{j, i}, \boldsymbol{w}_{j^{\prime}})\big)\Big]\Bigg) 
\tag{22a}
\label{18a}\\
& \leq \frac{1}{{n+m}} \sum_{j=1}^{n+m} \log \bigg(\sum_{j^{\prime}=1}^{n+m} \mathbb{E}_i\Big[\exp \big(\left\|\boldsymbol{s}_{j, i}-\boldsymbol{w}_j\right\|^2\notag\\
&\hspace{10em} -\left\|\boldsymbol{s}_{j, i}-\boldsymbol{w}_{j^{\prime}}\right\|^2\big)\Big]\bigg) \tag{22b}
\label{18b}\\
& =\frac{1}{{n+m}} \sum_{j=1}^{n+m} \log \Bigg(\sum_{j^{\prime}=1}^{n+m} \mathbb{E}_i\Big[\exp \big(2(\boldsymbol{w}_{j^{\prime}}-\boldsymbol{w}_j)^{\top} \boldsymbol{s}_{j, i}\notag\\
&\hspace{11em}+(\boldsymbol{w}_{j^{\prime}}^{\top} \boldsymbol{w}_{j^{\prime}}-\boldsymbol{w}_j^{\top} \boldsymbol{w}_j)\big)\Big]\Bigg) \notag\\
& \leq \frac{1}{{n+m}} \sum_{j=1}^{n+m} \log \Bigg(\sum_{j^{\prime}=1}^{n+m} \mathbb{E}_i\Big[\exp \big((\boldsymbol{w}_{j^{\prime}}-\boldsymbol{w}_j\big)^{\top} (\widehat{\boldsymbol{p}}_j\notag\\
&\hspace{5em}+\widehat{\boldsymbol{v}}_{j, i})+(\boldsymbol{w}_{j^{\prime}}^{\top} \boldsymbol{w}_{j^{\prime}}-\boldsymbol{w}_j^{\top} \boldsymbol{w}_j)\big)\Big]\Bigg) 
\tag{22c}
\label{18c}\\
& =\frac{1}{{n+m}} \sum_{j=1}^{n+m} \log \bigg(\sum_{j^{\prime}=1}^{n+m} \exp ((\boldsymbol{w}_{j^{\prime}}-\boldsymbol{w}_j)^{\top}(\widehat{\boldsymbol{p}}_j+\boldsymbol{\mu}_j)+ \notag\\
&(\boldsymbol{w}_{j^{\prime}}^{\top} \boldsymbol{w}_{j^{\prime}}-\boldsymbol{w}_j^{\top} \boldsymbol{w}_j)+\frac{1}{2}(\boldsymbol{w}_{j^{\prime}}-\boldsymbol{w}_j)^{\top} \boldsymbol{\Sigma}_j(\boldsymbol{w}_{j^{\prime}}-\boldsymbol{w}_j))\bigg)
\tag{22d}
\label{18d}\\
& =\frac{1}{{n+m}} \sum_{j=1}^{n+m}-\log \frac{\exp \Big(\boldsymbol{w}_jt(\widehat{\boldsymbol{p}}_j}{\exp \Big(\boldsymbol{w}_{j^{\prime}}(\widehat{\boldsymbol{p}}_j+\boldsymbol{\mu}_j)+(\boldsymbol{w}_{j^{\prime}}^{\top} \boldsymbol{w}_{j^{\prime}}} \notag \\
&\hspace{4em} \frac{+\boldsymbol{\mu}_j)\Big)}{-\boldsymbol{w}_j^{\top} \boldsymbol{w}_j)+\frac{1}{2}(\boldsymbol{w}_{j^{\prime}}-\boldsymbol{w}_j)^{\top} \boldsymbol{\Sigma}_j(\boldsymbol{w}_{j^{\prime}}-\boldsymbol{w}_j)\Big)}\notag\\
& =\overline{\mathcal{L}}_{\infty}(\boldsymbol{W}) \text {. } \notag\\
&\notag
\end{align}

Here, we introduce several key derivations. 
Eq.~\eqref{18a} follows from the Jensen's inequality $\mathbb{E}[\log (\boldsymbol{x})] \leq \log \mathbb{E}[\boldsymbol{x}]$. For Eq.~\eqref{18b}, we define a function $f=d^c(\boldsymbol{x}, \boldsymbol{y})-\|\boldsymbol{x}-\boldsymbol{y}\|^2$ and compute the derivative of $f$. We find $f$ is an increasing function when $\|\boldsymbol{x}-\boldsymbol{y}\|^2<\frac{\sqrt{1+|c|}-\sqrt{|c|}}{\sqrt{|c|}}$. In this case, if we have $\left\|\boldsymbol{s}_{j, i}-\boldsymbol{w}_j\right\|^2<\left\|\boldsymbol{s}_{j, i}-\boldsymbol{w}_{j^{\prime}}\right\|^2<\frac{\sqrt{1+|c|}-\sqrt{|c|}}{\sqrt{|c|}}$, we can derive $d^c\left(\boldsymbol{s}_{j, i}, \boldsymbol{w}_j\right)-\left\|\boldsymbol{s}_{j, i}, \boldsymbol{w}_j\right\|^2<d^c\left(\boldsymbol{s}_{j, i}, \boldsymbol{w}_{j^{\prime}}\right)-\left\|\boldsymbol{s}_{j, i}, \boldsymbol{w}_{j^{\prime}}\right\|^2 \quad$ and $\left\|\boldsymbol{s}_{j, i}, \boldsymbol{w}_j\right\|^2-\left\|\boldsymbol{s}_{j, i}, \boldsymbol{w}_{j^{\prime}}\right\|^2>d^c\left(\boldsymbol{s}_{j, i}, \boldsymbol{w}_j\right)-d^c\left(\boldsymbol{s}_{j, i}, \boldsymbol{w}_{j^{\prime}}\right)$. Eq.~\eqref{18c} is obtained by replacing $\boldsymbol{s}_{j, i}$ with the augmentation process using the parallel transport and exponential map, \emph{i.e.}, $\boldsymbol{s}_{j, i}=\operatorname{expm}_{\boldsymbol{p}_j}^c\left(\mathrm{PT}_{\mathbf{0} \rightarrow \boldsymbol{p}_j}^c\left(\widehat{\boldsymbol{v}}_{j, i}\right)\right)$, and adding scale constraints on $c,\left\|\widehat{\boldsymbol{v}}_{j, i}\right\|$, and $\left\|\widehat{\boldsymbol{p}}_j\right\|$. Eq.~\eqref{18d} is obtained by using the moment-generating function: $\mathbb{E}[\exp (\boldsymbol{x})]=\exp \left(\boldsymbol{\mu}+\frac{1}{2} \boldsymbol{\Sigma}\right), \boldsymbol{x} \sim \mathcal{N}(\boldsymbol{\mu}, \boldsymbol{\Sigma})$. In our derivation, $\left(\boldsymbol{w}_{j^{\prime}}-\boldsymbol{w}_j\right)^{\top}\left(\widehat{\boldsymbol{p}}_j+\widehat{\boldsymbol{v}}_{j, i}\right)+\left(\boldsymbol{w}_{j^{\prime}}^{\top} \boldsymbol{w}_{j^{\prime}}-\boldsymbol{w}_j^{\top} \boldsymbol{w}_j\right)$ is a Gaussian vector.

\subsection{Training}
\label{training}

In this work, our goal is to learn gradient flow networks $F_{1\sim 6}$ that can appropriately estimate $c$, $\boldsymbol{\mu}$, $\boldsymbol{L}$ in Eq.~\eqref{eq:ode} and $\delta_{\boldsymbol{p}}$, $\delta_{\boldsymbol{\mu}}$, $\delta_{\boldsymbol{L}}$ in Eq.~\eqref{eq:per}, ultimately facilitating the learning of $\boldsymbol{W}$.
To this end, 
we utilize meta-learning to learn $F_{1\sim 6}$ via a bi-level optimization manner. 
In the inner loop, we estimate the distributions of seen classes in training dataset $\mathcal{D}_t$ by using $F_1$, $F_2$ and $F_3$, and distributions of unseen classes beyond $\mathcal{D}_t$ by using $F_4$, $F_5$ and $F_6$. Then, we use the Riemannian gradient descent algorithm~\citep{absil2008optimization} to iteratively train the classifier $\boldsymbol{W}$ by minimizing the upper bound $\overline{\mathcal{L}}_{\infty}(\boldsymbol{W})$ in Eq.~\eqref{loss:inifite11}, and obtain the updated classifier $\boldsymbol{W}^*=\left[\boldsymbol{w}_1^*, \cdots, \boldsymbol{w}_{n+m}^*\right]$. In the outer loop, we update the gradient flow networks $F_{1\sim 6}$ by minimizing the following meta-objective that is the classification loss of $\boldsymbol{W}^*$ on the validation data $\mathcal{D}_v$ along with the proposed hierarchy-preserving regularization loss,
\begin{equation}\small
    \begin{aligned}
&\min _{F_1, F_2, F_3, F_4, F_5, F_6} \mathcal{L}\left(\boldsymbol{W}^*\right)=  \mathbb{E}_{\boldsymbol{x} \sim \mathcal{D}_v}\left[-\sum_{j=1}^{n+m} \mathds{1}_{y=j} \log \right.\\
&\left. \frac{\exp \Big(-d^c(\boldsymbol{x}, \boldsymbol{w}_j^*)\Big)}{\sum_{j^{\prime}=1}^{n+m} \exp \left(-d^c\left(\boldsymbol{x}, \boldsymbol{w}_{j^{\prime}}^*\right)\right)}\right]+\beta R_{Hier} \\ 
&\text { s.t. } \boldsymbol{W}^*=\arg \min _{\boldsymbol{W}} \overline{\mathcal{L}}_{\infty}(\boldsymbol{W}).
\end{aligned}
\tag{23}
\label{loss:outer}
\end{equation}
Here, $\beta$ is a hyperparameter that controls the strength of the hierarchy-preserving regularization loss. $\mathds{1}_{y=j}$ is the indicator function, meaning that if the label $y$ of $\boldsymbol{x}$ is equal to $j, \mathds{1}_{y=j}=1$, and 0 otherwise. The pseudo code of training is summarized in Algorithm~\ref{alg:train}.

\section{Experiments}\label{experi}
In this section, we conduct comprehensive experiments
to verify the superiority of the proposed hyperbolic dual feature augmentation (HDFA) method. We evaluate the HDFA method on five tasks, including class-incremental learning, few-shot open-set recognition, few-shot learning, zero-shot learning, and general image classification tasks. Then, we conduct ablation, visualization and efficiency experiments to demonstrate the effectiveness of dual feature augmentation in the hyperbolic space. 

\subsection{Class-Incremental Learning}
\subsubsection{Task Setting}
Class incremental learning (CIL) trains a model in an online manner given a sequence of tasks, which is denoted by $T_{1: t} \equiv\left\{T_1, \cdots, T_t, \cdots, T_t\right\}$. Each task $T_t$ is defined by a training dataset $\mathcal{D}_t$ composed of examples with task-specific labels $y \in \mathcal{Y}_t$, where $\left(\mathcal{Y}_1 \cup \cdots \cup \mathcal{Y}_{t-1}\right) \cap \mathcal{Y}_t=\emptyset$. At the $t^{\text {th }}$ incremental stage, the model is trained on $\mathcal{D}_t^{\prime}=$ $\mathcal{D}_t \cup \mathcal{B}_{t-1}$, where $\mathcal{B}_{t-1}$ is a small subset of all previously seen training datasets, which is called a memory buffer. The performance of the trained model is evaluated on the test data sampled from a collection of all the encountered tasks without task boundaries.

\subsubsection{Experimental Setup}
\textit{Datasets.} We evaluate our method on the CIFAR-
100~\citep{krizhevsky2009learning} and Tiny-ImageNet~\citep{le2015tiny} datasets that have hyperbolic structures and are commonly used in class-incremental learning. CIFAR-100 contains 60,000 images of 32 × 32 size from 100 classes, and each class includes 500 training images and 100 test images. Tiny-ImageNet contains 200 classes, and each class contains 500 training images, 50 validation images, and 50 test images. We split the CIFAR-100 dataset in three different settings: C100-B50-S5, C100-B50-S10 and C100-B50-S20. For instance, C100-B50-S5 represents that we first pre-train the model using the first 50 classes, and the following classes are split into 5 stages each has 10 classes. Similarly, the settings for Tiny-ImageNet are T200-B100-S5, T200-B100-S10, and T200-B100-S20.

\noindent\textit{Baselines.} We compare HDFA with the following methods: LwF-MC~\citep{rebuffi2017icarl}, iCaRL~\citep{rebuffi2017icarl}, EEIL~\citep{castro2018end}, MAS~\citep{aljundi2018memory}, LwM~\citep{dhar2019learning}, LUCIR~\citep{hou2019learning}, MUC~\citep{liu2020more}, SDC~\citep{yu2020semantic}, ABD\citep{smith2021always}, IL2A~\citep{zhu2021class}, PASS~\citep{zhu2021prototype}, R-DFCIL~\citep{gao2022r}, HFA~\citep{gao2022hyperbolic},  ${\rm FeTrIL}_{fc}^{1}$~\citep{petit2023fetril} and EDG~\citep{gao2023exploring}.

\begin{figure}[t]
    \centering
    \includegraphics[width=1\columnwidth]{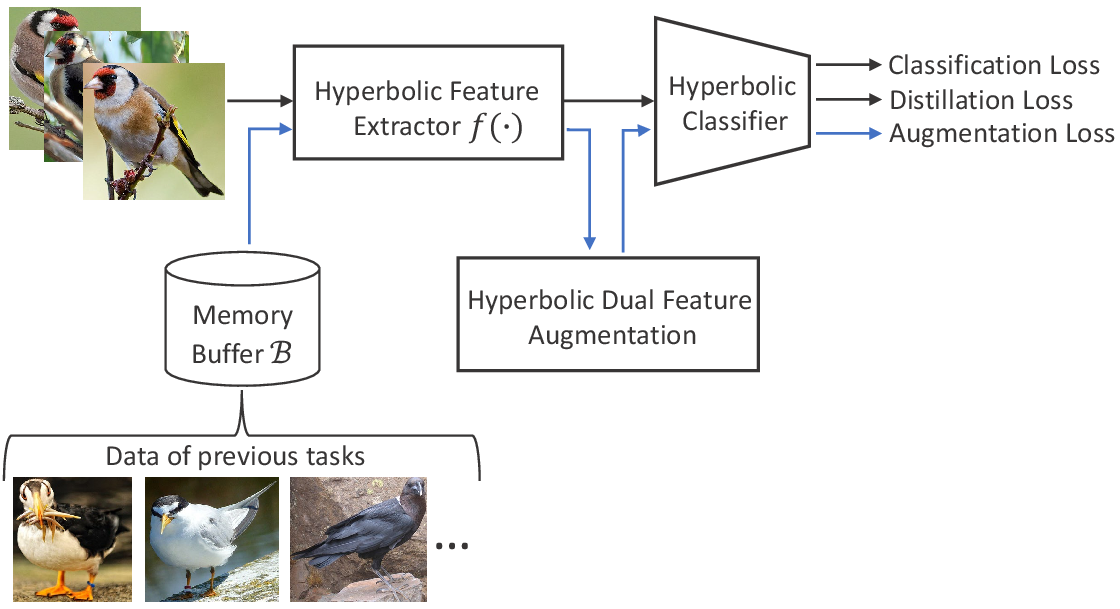}
    \caption{The overview of the proposed HDFA method in the class-incremental learning task.}
    \label{CIL_aug}
\end{figure}

\begin{table*}[t]
    \centering

    \caption{Final accuracy (\%) on the CIFAR100 dataset with different numbers of incremental stages. ‘Euclidean’ and ‘Hyperbolic’ denote the algorithms are in Euclidean spaces and hyperbolic spaces, respectively.}
    
    \scalebox{0.85}{\begin{tabular}{lcc}
        \hline
         Methods & Category &\makecell[c]{CIFAR100 \\ C100-B50-S5\hspace{0.4cm} C100-B50-S10 \hspace{0.4cm} C100-B40-S20}  \\
         \toprule
     	\specialrule{0em} {1.5pt}{1.5pt}
         MAS~\citep{aljundi2018memory} & Euclidean &
         \makecell[c]{13.32\hspace{2cm}11.32\hspace{2.2cm}8.56}\\
         \specialrule{0em} {1.5pt}{1.5pt}
         LwF-MC~\citep{rebuffi2017icarl} &Euclidean &
         \makecell[c]{33.47\hspace{2cm}15.85\hspace{2cm}12.00}\\
         \specialrule{0em} {1.5pt}{1.5pt}
         LwM~\citep{dhar2019learning} &Euclidean &
         \makecell[c]{35.67\hspace{2cm}20.64\hspace{2cm}12.70}\\
         \specialrule{0em} {1.5pt}{1.5pt}
         MUC~\citep{liu2020more} &Euclidean &
         \makecell[c]{37.26\hspace{2cm}18.64\hspace{2cm}13.01}\\
         \specialrule{0em} {1.5pt}{1.5pt}
         iCaRLCNN~\citep{rebuffi2017icarl} &Euclidean &
         \makecell[c]{40.19\hspace{2cm}38.87\hspace{2cm}34.26}\\
     	\specialrule{0em} {1.5pt}{1.5pt}
         iCaRLNME~\citep{rebuffi2017icarl} & Euclidean &\makecell[c]{49.14\hspace{2cm}45.31\hspace{2cm}40.53} \\
         \specialrule{0em} {1.5pt}{1.5pt}
         EEIL~\citep{castro2018end} &Euclidean & \makecell[c]{50.21\hspace{2cm}47.60\hspace{2cm}42.23} \\
     	\specialrule{0em} {1.5pt}{1.5pt}
         LUCIR~\citep{hou2019learning} &Euclidean & \makecell[c]{54.71\hspace{2cm}50.53\hspace{2cm}48.00} \\
         \specialrule{0em} {1.5pt}{1.5pt}
         IL2A~\citep{zhu2021class} & Euclidean &\makecell[c]{54.98\hspace{2cm}45.07\hspace{2cm}45.74} \\
         \specialrule{0em} {1.5pt}{1.5pt}
         R-DFCIL~\citep{gao2022r} & Euclidean &\makecell[c]{54.76\hspace{2cm}49.70\qquad\hspace{2cm}-} \\
         \specialrule{0em} {1.5pt}{1.5pt}
         HFA~\citep{gao2022hyperbolic}  & Hyperbolic &\makecell[c]{55.55\hspace{2cm}52.11\hspace{2cm}46.12}\\
     	\specialrule{0em} {1.5pt}{1.5pt}
         \textbf{HDFA (Ours)} & Hyperbolic &\makecell[c]{\textbf{56.62}\hspace{1.85cm}\textbf{54.42}\hspace{1.9cm}\textbf{48.38}}\\
         \hline
    \end{tabular}}
    \label{tab:final_cifar100}
\end{table*}
\begin{table*}[t]
    \centering

    \caption{Final accuracy (\%) on the Tiny-ImageNet dataset with different numbers of incremental stages. `Euclidean’, `Hyperbolic’, and `Mixed-curvature' denote the algorithms are in Euclidean spaces, hyperbolic spaces, and mixed-curvature spaces, respectively.}
    \scalebox{0.85}{\begin{tabular}{lcc}
        \hline
         Methods & Category&\makecell[c]{Tiny-ImageNet \\ T200-B100-S5\hspace{0.4cm} T200-B100-S10 \hspace{0.4cm} T200-B100-S20}  \\
         \toprule
         \specialrule{0em} {1.5pt}{1.5pt}
         MAS~\citep{aljundi2018memory} &Euclidean&
         \makecell[c]{11.64\hspace{2.2cm}6.56\hspace{2.2cm}3.23}\\
         \specialrule{0em} {1.5pt}{1.5pt}
         LwF-MC~\citep{rebuffi2017icarl} &Euclidean&
         \makecell[c]{15.56\hspace{2cm}13.26\hspace{2.2cm}7.87}\\
         \specialrule{0em} {1.5pt}{1.5pt}
         MUC~\citep{liu2020more} &Euclidean&
         \makecell[c]{17.23\hspace{2cm}15.21\hspace{2.2cm}8.26}\\
         \specialrule{0em} {1.5pt}{1.5pt}
         iCaRLCNN~\citep{rebuffi2017icarl} &Euclidean&
         \makecell[c]{23.17\hspace{2cm}20.71\hspace{2cm}20.28}\\
     	\specialrule{0em} {1.5pt}{1.5pt}
         iCaRLNME~\citep{rebuffi2017icarl} &Euclidean&\makecell[c]{34.43\hspace{2cm}33.24\hspace{2cm}27.51} \\
         \specialrule{0em} {1.5pt}{1.5pt}
         EEIL~\citep{castro2018end} &Euclidean& \makecell[c]{35.00\hspace{2cm}33.67\hspace{2cm}27.64} \\
         \specialrule{0em} {1.5pt}{1.5pt}
         IL2A~\citep{zhu2021class} &Euclidean& \makecell[c]{36.58\hspace{2cm}34.28\hspace{2cm}28.34} \\
         \specialrule{0em} {1.5pt}{1.5pt}
         EDG~\citep{gao2023exploring} &Mixed-curvature& \makecell[c]{38.10\hspace{1.95cm}\textbf{37.99}\hspace{1.95cm}34.85} \\
         \specialrule{0em} {1.5pt}{1.5pt}
         HFA~\citep{gao2022hyperbolic}  &Hyperbolic&\makecell[c]{36.11\hspace{2cm}33.65\hspace{2cm}33.73}\\
     	\specialrule{0em} {1.5pt}{1.5pt}
         \textbf{HDFA (Ours)} &Hyperbolic& \makecell[c]{\textbf{38.23}\hspace{1.95cm}36.34\hspace{1.95cm}\textbf{35.15}}\\
         \hline
    \end{tabular}}
    \label{tab:final_tiny}
\end{table*}

\noindent\textit{Implementation details.} In class-incremental learning, reply-based methods have shown effectiveness by storing data ~\citep{chaudhry2018efficient,wu2019large,douillard2020podnet}, yet they cannot store much data due to memory limitation.
Here, we use HDFA to perform augmentation for the limited stored data. Figure~\ref{CIL_aug} shows the pipeline of HDFA in the class-incremental learning task. The model contains a hyperbolic feature extractor $f(\cdot)$, a hyperbolic distance-based classifier $C(\cdot)$, a memory buffer $\mathcal{B}$, and the proposed hyperbolic dual feature augmentation module containing six
gradient flow networks $F_{1}$, $F_{2}$, $F_{3}$, $F_{4}$, $F_{5}$ and $F_{6}$. We use ResNet18~\citep{he2016deep} as the feature extractor $f(\cdot)$. Features extracted from $f(\cdot)$ are considered to be located in the tangent space at the origin and then projected to the hyperbolic space via an exponential map. Following existing reply-based continual learning methods~\citep{rebuffi2017icarl,castro2018end, hou2019learning}, we use the memory buffer $\mathcal{B}$ to store 20 samples for each class and perform HDFA for the stored data.

\begin{table*}[t]
    \centering

    \caption{Average incremental accuracy (\%) on the CIFAR100 dataset with different numbers of incremental stages. ‘Euclidean’ and ‘Hyperbolic’ denote the algorithms are in Euclidean spaces and hyperbolic spaces, respectively.}
    \scalebox{0.85}{\begin{tabular}{lcc}
        \hline
         Methods & Category&\makecell[c]{CIFAR100 \\ C100-B50-S5\hspace{0.4cm} C100-B50-S10 \hspace{0.4cm} C100-B40-S20}  \\
         \toprule
         
         iCaRLCNN~\citep{rebuffi2017icarl} &Euclidean &
         \makecell[c]{51.31\hspace{2cm}48.28\hspace{2cm}44.61}\\
     	\specialrule{0em} {1.5pt}{1.5pt}
         iCaRLNME~\citep{rebuffi2017icarl} & Euclidean &\makecell[c]{60.86\hspace{2cm}53.86\hspace{2cm}51.01} \\
         \specialrule{0em} {1.5pt}{1.5pt}
         EEIL~\citep{castro2018end} &Euclidean & \makecell[c]{60.40\hspace{2cm}55.77\hspace{2cm}52.57} \\
     	\specialrule{0em} {1.5pt}{1.5pt}
         LUCIR~\citep{hou2019learning} &Euclidean & \makecell[c]{63.75\hspace{2cm}60.68\hspace{2cm}58.27} \\
         \specialrule{0em} {1.5pt}{1.5pt}
         SDC~\citep{yu2020semantic} &Euclidean & \makecell[c]{56.90\hspace{2cm}57.00\hspace{2cm}58.90} \\
         \specialrule{0em} {1.5pt}{1.5pt}
         ABD~\citep{smith2021always} &Euclidean & \makecell[c]{63.80\hspace{2cm}62.50\hspace{2cm}57.40} \\
         \specialrule{0em} {1.5pt}{1.5pt}
         IL2A~\citep{zhu2021class} &Euclidean & \makecell[c]{66.19\hspace{2cm}58.20\hspace{2cm}58.01} \\
         \specialrule{0em} {1.5pt}{1.5pt}
         PASS~\citep{zhu2021prototype} &Euclidean & \makecell[c]{63.84\hspace{2cm}59.87\hspace{2cm}58.07} \\
         \specialrule{0em} {1.5pt}{1.5pt}
         R-DFCIL~\citep{gao2022r} &Euclidean & \makecell[c]{64.78\hspace{2cm}61.71\hspace{2.6cm}  -  } \\
         \specialrule{0em} {1.5pt}{1.5pt}
         ${\rm FeTrIL}_{fc}^{1}$~\citep{petit2023fetril} & Euclidean &\makecell[c]{64.70\hspace{2cm}63.40\hspace{2cm}57.40} \\
         \specialrule{0em} {1.5pt}{1.5pt}
         
         \textbf{HDFA (Ours)} &Hyperbolic & \makecell[c]{\textbf{66.74}\hspace{1.95cm}\textbf{64.57}\hspace{1.85cm}\textbf{60.72}}\\
         \hline
    \end{tabular}}
    \label{tab:average_cifar100}
\end{table*}
In the training phase, a sequence of $t$ tasks is given with data $D^0, D^1, \cdots, D^t$. The training process of HDFA in class-incremental learning includes three steps. \begin{enumerate}
    \item[1)] In the first step, we use $D^0$ to train the hyperbolic feature extractor $f(\cdot)$ and the hyperbolic classifier $C(\cdot)$ using the following cross-entropy loss,
    \begin{equation}\small
    \mathcal{L}_c^0=\mathbb{E}_{\boldsymbol{I}^0 \in D^0}\bigg[-\sum_{j=1}^n \mathds{1}_{y=j} \log \frac{\exp \big(C\big(f(\boldsymbol{I}^0)\big)\big)}{\sum_{j^{\prime}=1}^n \exp \big(C\big(f(\boldsymbol{I}^0)\big)\big)}\bigg],
    \tag{24}
    \end{equation}
    where $\boldsymbol{I}^{0}\in D^0$ denotes the image with from the base data $D^0$ and $y$ denotes its label.
    \item[2)] In the second step, we train our hyperbolic dual feature augmentation module (\emph{i.e.}, gradient flow networks $F_{1}$, $F_{2}$, $F_{3}$, $F_{4}$, $F_{5}$ and $F_{6}$) via the bi-level optimization process presented in Sect.~\ref{training}. In each inner loop, $D^0$ is randomly split into $D_{t}^0$ and $D_{v}^0$ for training and validation, respectively. We estimate the distributions for $n^{0}$ seen classes in $D_{t}^0$ by using $F_{1}$, $F_{2}$, $F_{3}$, and estimate the distributions for $m^{0}$ synthesized unseen classes by using $F_{4}$, $F_{5}$ and $F_{6}$. Then, we use the mean of the $n^{0}+m^{0}$ classes to initialize the classifier and update it by the derived upper bound $ \overline{\mathcal{L}}_{\infty}$ of the augmentation loss. In the outer loop, we update the hyperbolic dual feature augmentation module by minimizing the loss in Eq.~\eqref{loss:outer}.
    \item[3)] In the third step, we train the model in the following tasks. Specifically, we train $f(\cdot)$ and $C(\cdot)$ by minimizing a cross-entropy classification loss $\mathcal{L}_c$ on data $D^{t}$ of the current task, along with a distillation loss $\mathcal{L}_d$ and the derived upper bound $ \overline{\mathcal{L}}_{\infty}$ of the augmentation loss on data (\emph{i.e.}, features of seen and unseen classes) of previous tasks stored in $\mathcal{B}$. The distillation loss $\mathcal{L}_d$ is defined as
\begin{equation}
    \mathcal{L}_d=\mathbb{E}_{\boldsymbol{I} \in \mathcal{B}}\left[d^c\left(f^{\prime}(\boldsymbol{I}), f(\boldsymbol{I})\right)\right],
    \tag{25}
\end{equation}
where $f^{\prime}(\cdot)$ denotes the feature extractor of the previous task.
\end{enumerate}
\noindent\textit{Evaluation metrics.} We use the final accuracy and average incremental accuracy~\citep{castro2018end,zhu2021class} to evaluate our method. We also provide per-state accuracy curves for a more detailed view of the accuracy evolution during the class-incremental learning process. 

\noindent\textit{Hyperparameters.} The ratio of the training set $\mathcal{D}_t$ to the validation set $\mathcal{D}_v$ is 1:1. The initial value of the curvature $c$ is set as -1. We employ the SGD optimizer in the three aforementioned training stages. In the first stage, we train the model for 100 epochs with a learning rate of 0.001. The second stage consists of 100 epochs with a learning rate of 0.01. The third stage consists of 5 epochs with a learning rate of 0.0003. The weight decay for the backbone is set as 0.0005. We store 20 samples per class in the memory module $\mathcal{B}$. In the second stage, 16 samples are used for training and another 16 samples for validation. The hyperparameter $\beta$ and $\gamma$ for the hierarchy-preserving regularization loss are set as 10 and 0.1, respectively. In the third stage, we sample 5 samples per class from $\mathcal{B}$ to estimate the distribution. The trade-off parameters for the augmentation loss $ \overline{\mathcal{L}}_{\infty}$ and the distillation loss $\mathcal{L}_d$ in the third stage are both set as 10.
\begin{figure*}[t]
    \centering
    \begin{subfigure}[b]{0.55\columnwidth}
        \centering
        \includegraphics[width=1.12\columnwidth]{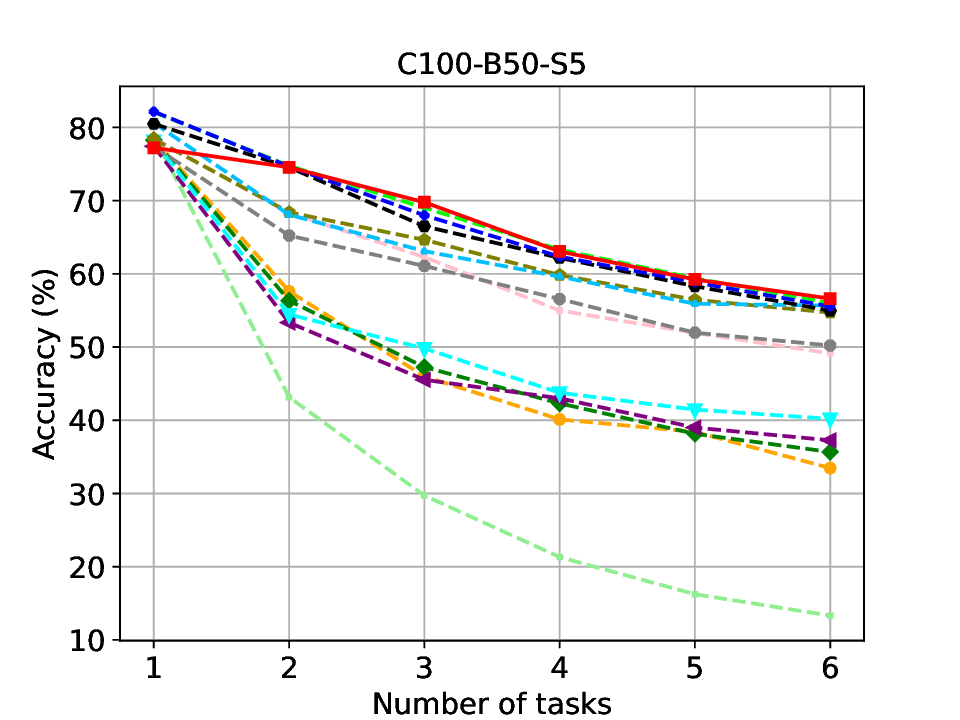}
    \end{subfigure}
    \begin{subfigure}[b]{0.55\columnwidth}
        \centering
        \includegraphics[width=1.12\columnwidth]{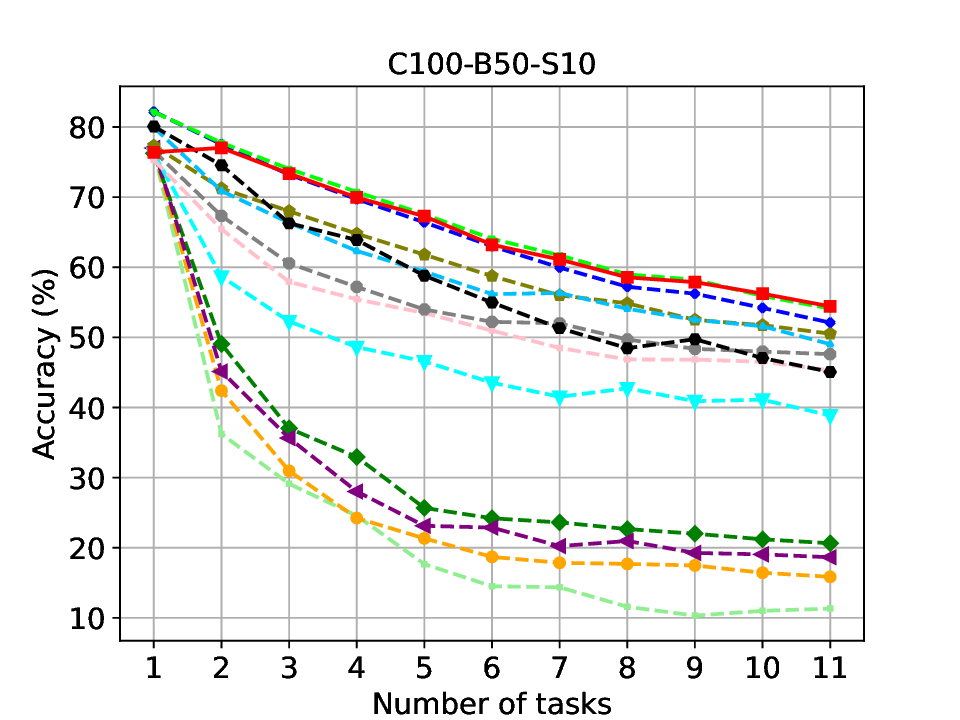}
    \end{subfigure}
        \begin{subfigure}[b]{0.55\columnwidth}
        \centering
        \includegraphics[width=1.12\columnwidth]{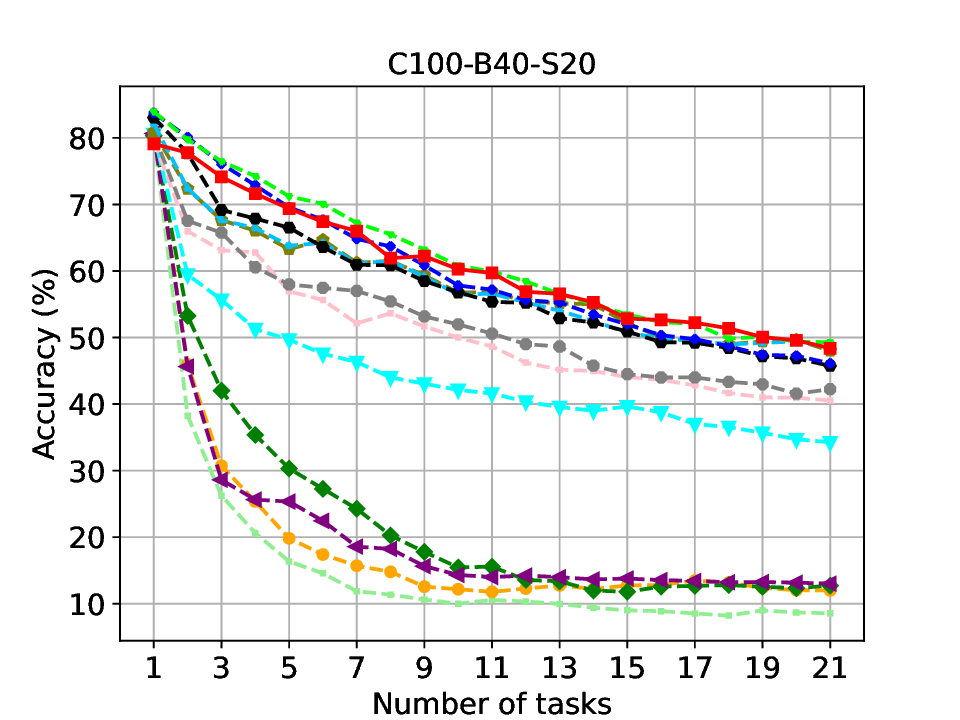}
    \end{subfigure}
            \begin{subfigure}[b]{0.2\columnwidth}
        \centering
        \includegraphics[width=1\columnwidth]{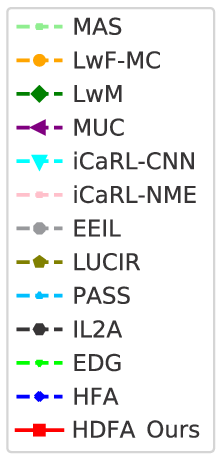}
    \end{subfigure}
    \begin{subfigure}[b]{0.55\columnwidth}
        \centering
        \includegraphics[width=1.12\columnwidth]{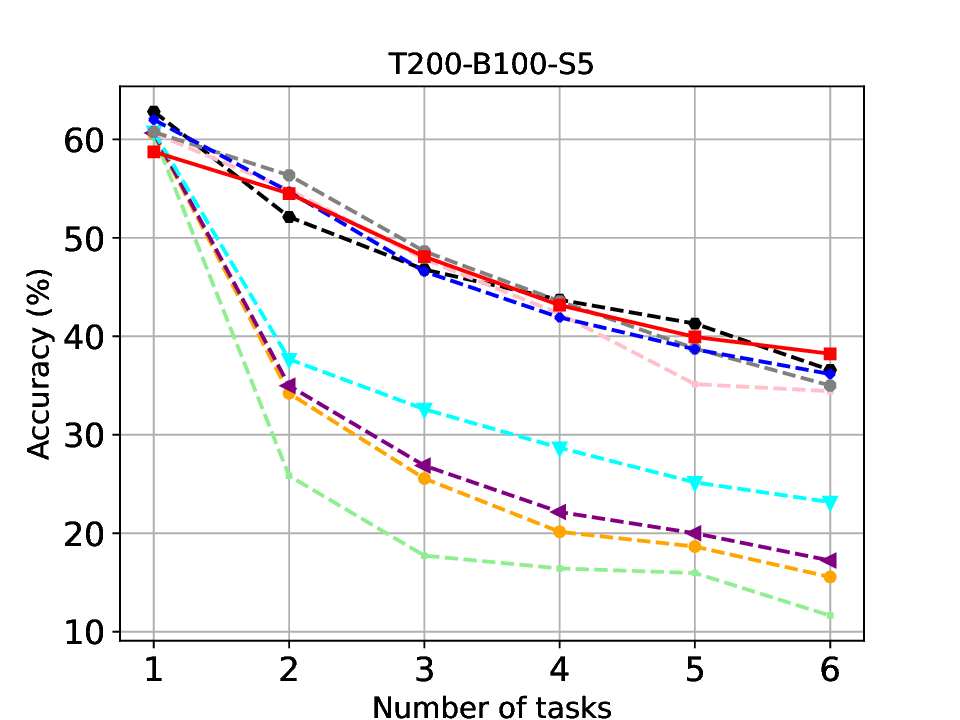}
    \end{subfigure}
    \begin{subfigure}[b]{0.55\columnwidth}
        \centering
        \includegraphics[width=1.12\columnwidth]{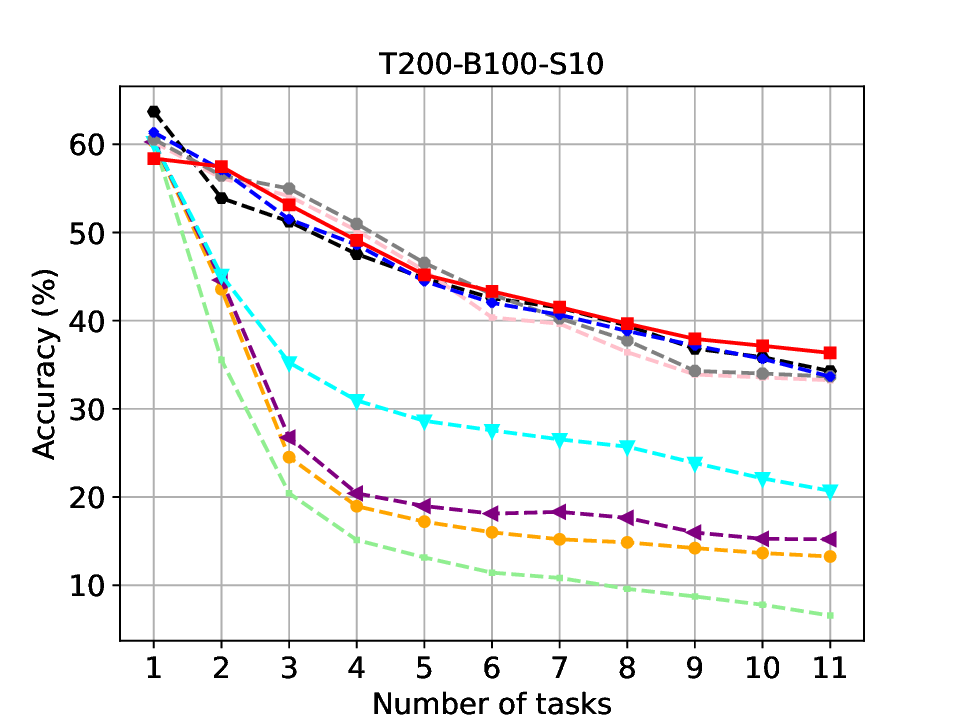}
    \end{subfigure}
        \begin{subfigure}[b]{0.55\columnwidth}
        \centering
        \includegraphics[width=1.12\columnwidth]{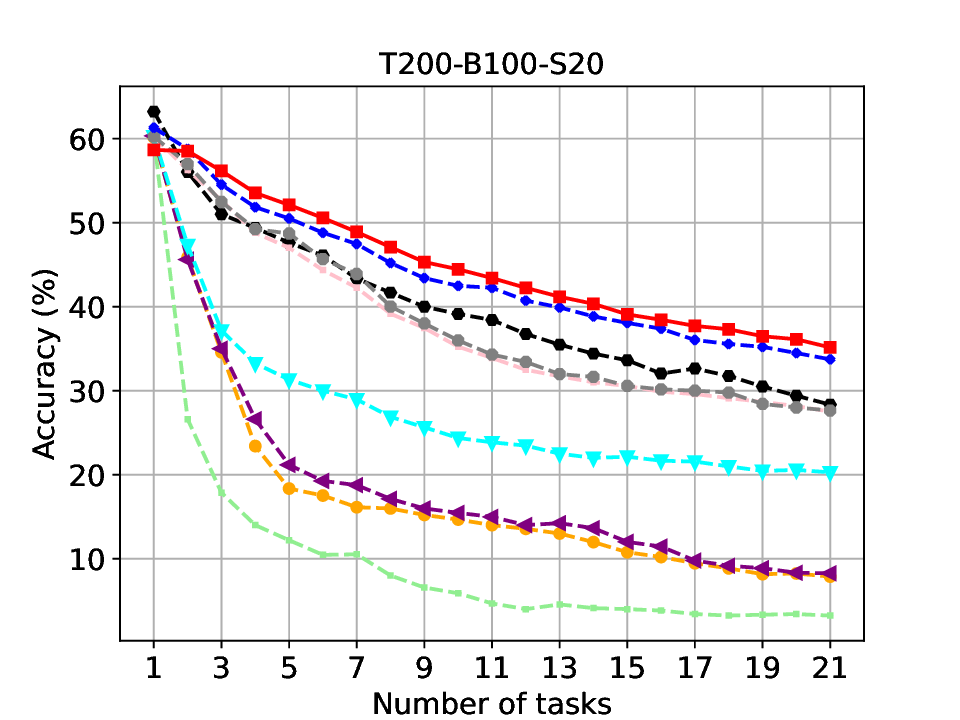}
    \end{subfigure}
            \begin{subfigure}[b]{0.2\columnwidth}
        \centering
        \includegraphics[width=1\columnwidth]{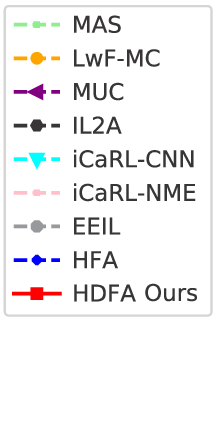}
    \end{subfigure}

    \caption{Curves of top-1 accuracy on the CIFAR-100 and Tiny-ImageNet datasets under different settings.}
    \label{CIL_image}

\end{figure*}

\subsubsection{Results}
Final accuracies on the C100-B50-S5, C100-B50-S10,
and C100-B40-S20 settings are shown in Table~\ref{tab:final_cifar100}. Final accuracies on the T200-B100-S5, T200-B100-S10, and T200-B100-S20 settings are shown in Table~\ref{tab:final_tiny}. Average incremental accuracies on the C100-B50-S5, C100-B50-S10, and C100-B40-S20 settings are shown in Table~\ref{tab:average_cifar100}. Accuracy curves
on the C100-B50-S5, C100-B50-S10, and C100-B40-S20
settings are shown in Figure~\ref{CIL_image}. We observe that our method achieves better performance than the representative data replay-based methods, such as iCaRL~\citep{rebuffi2017icarl}, EEIL~\citep{castro2018end} and LUCIR~\citep{hou2019learning}. This suggests that our method can well estimate the distribution from limited data and perform diverse and discriminative augmentation for the stored data to improve class incremental learning performance. When compared to the augmentation-based method IL2A~\citep{zhu2021class}, our method remarkably shows stronger performance. This demonstrates the effectiveness of modeling data in the hyperbolic space, which better captures the hierarchical geometry implied in data. When compared to the EDG method~\citep{gao2023exploring} that uses the mixed-curvature space to embed data, our method exceeds it in most settings and performs competitively in certain settings. 
Since EDG does not augment data, the results show the effectiveness of our hyperbolic feature augmentation method.
Particularly, we observe that the proposed HDFA has better performance than HFA that only performs feature augmentation for the seen classes in the hyperbolic space. The reason is that augmentation for both seen and unseen classes prepares the model for future classes and enables it to learn more transferable and diverse representations, and in turn, reduces the forgetting of old classes when learning new classes.

\begin{table*}[t]
    \centering

    \caption{Accuracy (\%) comparisons with few-shot open-set recognition methods on the mini-ImageNet dataset. ‘Euclidean’ and ‘Hyperbolic’ denote the algorithms are in Euclidean spaces and hyperbolic spaces, respectively.}
    \renewcommand{\arraystretch}{1.3}
    \scalebox{0.9}{\begin{tabular}{lccc}
        \hline
         Methods &Category &\makecell[c]{1-shot \\ ACC\hspace{1.2cm} AUROC} & \makecell[c]{5-shot \\ ACC\hspace{1.2cm} AUROC} \\
         \hline
        \makecell[l]{ProtoNet~\citep{snell2017prototypical}\\FEAT~\citep{ye2020few}\\PEELER~\citep{liu2020few}\\SnaTCHer-F~\citep{jeong2021few}\\SnaTCHer-T~\citep{jeong2021few}\\SnaTCHer-L~\citep{jeong2021few}\\ATT~\citep{huang2022task}\\ATT-G~\citep{huang2022task}\\RFDNet~\citep{deng2022learning}\\HFA~\citep{gao2022hyperbolic}\\REFOCS~\citep{nag2023reconstruction}\\GEL~\citep{wang2023glocal}\\OPP~\citep{sun2024overall}\\ \textbf{HDFA (Ours)}}&\makecell[c]{Euclidean\\Euclidean\\Euclidean\\Euclidean\\Euclidean\\Euclidean\\Euclidean\\Euclidean\\Euclidean\\Hyperbolic\\Euclidean\\Euclidean\\Euclidean\\Hyperbolic}&\makecell[c]{64.01 $\pm$ 0.88\hspace{0.4cm} 51.81 $\pm$ 0.93\\67.02 $\pm$ 0.85 \hspace{0.4cm}57.01 $\pm$ 0.84\\65.86 $\pm$ 0.85\hspace{0.4cm} 60.57 $\pm$ 0.83\\67.02 $\pm$ 0.85 \hspace{0.4cm}68.27 $\pm$ 0.96\\66.60 $\pm$ 0.80\hspace{0.4cm} 70.17 $\pm$ 0.88\\67.60 $\pm$ 0.83\hspace{0.4cm} 69.40 $\pm$ 0.92\\67.64 $\pm$ 0.81\hspace{0.4cm} 71.35 $\pm$ 0.68\\68.11 $\pm$ 0.81\hspace{0.4cm} 72.41 $\pm$ 0.72\\66.23 $\pm$ 0.80 \hspace{0.4cm}71.37 $\pm$ 0.80\\69.37 $\pm$ 0.88\hspace{0.4cm} 74.38 $\pm$ 0.72 \\66.35 $\pm$ 0.87\hspace{0.4cm} 70.68 $\pm$ 0.60 \\68.26 $\pm$ 0.85\hspace{0.4cm} 73.70 $\pm$ 0.82\\68.52 $\pm$ 0.70\hspace{0.4cm} 72.68 $\pm$ 0.68\\\textbf{70.52 $\pm$ 0.85}\hspace{0.15cm} \textbf{76.53 $\pm$ 0.73}}&\makecell[l]{80.09 $\pm$ 0.58\hspace{0.4cm} 60.39 $\pm$ 0.92\\82.02 $\pm$ 0.53\hspace{0.4cm} 63.18 $\pm$ 0.78\\80.61 $\pm$ 0.59\hspace{0.4cm} 67.35 $\pm$ 0.80\\82.02 $\pm$ 0.53\hspace{0.4cm} 77.42 $\pm$ 0.73\\81.77 $\pm$ 0.53\hspace{0.4cm} 76.66 $\pm$ 0.78\\82.36 $\pm$ 0.58\hspace{0.4cm} 76.15 $\pm$ 0.83\\82.31 $\pm$ 0.49\hspace{0.4cm} 79.85 $\pm$ 0.58\\83.12 $\pm$ 0.48\hspace{0.4cm} 79.85 $\pm$ 0.57\\82.44 $\pm$ 0.54\hspace{0.4cm} 80.31 $\pm$ 0.59\\82.31 $\pm$ 0.59\hspace{0.4cm} 81.51 $\pm$ 0.60 \\82.44 $\pm$ 0.98\hspace{0.4cm} 76.83 $\pm$ 0.60\\83.05 $\pm$ 0.55\hspace{0.4cm} 82.29 $\pm$ 0.60\\ \textbf{83.57 $\pm$ 0.55}\hspace{0.15cm} 81.76 $\pm$ 0.60\\82.91 $\pm$ 0.62\hspace{0.4cm}\textbf{84.59 $\pm$ 0.55}} \\
         \hline
    \end{tabular}}
    \label{tab:fsor}
\end{table*}

\subsection{Few-Shot Open-Set Recognition}
\subsubsection{Task Setting}
Few-shot open-set recognition (FSOR) aims to classify seen classes and detect unseen ones with limited training samples. Formally, for a N-way K-shot Q-query problem sampled from dataset $\mathcal{D}$, the FSOR task can be presented as: $\mathcal{T}=\left\{\mathcal{S}, \mathcal{Q}_k, \mathcal{Q}_u\right\}$, where $\mathcal{S}=\left\{x_i, y_i\right\}_{i=1}^{|\mathcal{S}|}$ and $\mathcal{Q}_k=\left\{x_i, y_i\right\}_{i=1}^{\left|\mathcal{Q}_k\right|}$ are support set and seen query set, respectively. The label $y$ corresponding to the image $x$ in these two sets are from closed-set categories $\mathcal{C}_k$. Different from few shot learning tasks, $\mathcal{Q}_u=\left\{x_i, y_i\right\}_{i=1}^{\left|\mathcal{Q}_u\right|}$ is a set of query samples from unseen classes $\mathcal{C}_u$, and $\mathcal{C}_k \cap \mathcal{C}_u=\varnothing$. The goal of FSOR is to train a model with small support set $\mathcal{S}$ that can classify known queries $\mathcal{Q}_k$ and detect unknown queries $\mathcal{Q}_u$.
\subsubsection{Experimental Setup}
\textit{Datasets.} In FSOR, we evaluate our method on the mini-ImageNet~\citep{vinyals2016matching} and CIFAR-FS~\citep{krizhevsky2009learning} datasets. The mini-ImageNet dataset consists of 100 classes from ImageNet~\citep{b54} with 600 images for each class, involving 64 training classes, 16 validation classes, and 20 test classes. The CIFAR-FS dataset is the subset of CIFAR100~\citep{krizhevsky2009learning}. It contains a total of
60, 000 images of size 32 × 32 in 100 categories. The categories are divided in the same way as mini-ImageNet.

\noindent\textit{Baselines.} We compare our HDFA with the few-shot learning approaches
ProtoNet~\citep{snell2017prototypical} and FEAT~\citep{ye2020few}. 
Considering most of the existing FSOR methods are metric-based methods, ProtoNet and FEAT are representative and competitive methods.
In order to evaluate the open-set recognition performance, we input the open-set samples into their closed-set classifier
and recognize them by thresholding the entropy. We also
make a comparison with popular FSOR approaches PEELER~\citep{liu2020few}, SnaTCHer~\citep{jeong2021few}, ATT~\citep{huang2022task}, RFDNet~\citep{deng2022learning}, GEL~\citep{wang2023glocal}, REFOCS~\citep{nag2023reconstruction} and OPP~\citep{sun2024overall}. 

\begin{table*}[t]
    \centering

    \caption{Accuracy (\%) comparisons with few-shot open-set recognition methods on the CIFAR-FS dataset. ‘Euclidean’ and ‘Hyperbolic’ denote the algorithms are in Euclidean spaces and hyperbolic spaces, respectively.}
    \renewcommand{\arraystretch}{1.3}
    \scalebox{0.9}{\begin{tabular}{lccc}
        \hline
         Methods & Category&\makecell[c]{1-shot \\ ACC\hspace{1.2cm} AUROC} & \makecell[c]{5-shot \\ ACC\hspace{1.2cm} AUROC} \\
         \hline
        \makecell[l]{ProtoNet~\citep{snell2017prototypical}\\FEAT~\citep{ye2020few}\\PEELER~\citep{liu2020few}\\SnaTCHer-F~\citep{jeong2021few}\\HFA~\citep{gao2022hyperbolic}\\REFOCS~\citep{nag2023reconstruction}\\ \textbf{HDFA (Ours)}}&\makecell[c]{Euclidean\\Euclidean\\Euclidean\\Euclidean\\Hyperbolic\\Euclidean\\Hyperbolic}&\makecell[c]{66.01 $\pm$ 0.57\hspace{0.4cm} 66.79 $\pm$ 0.80\\70.89 $\pm$ 0.89\hspace{0.4cm} 74.83 $\pm$ 0.79\\71.47 $\pm$ 0.67\hspace{0.4cm} 71.28 $\pm$ 0.57\\67.02 $\pm$ 0.89\hspace{0.4cm} 73.94 $\pm$ 0.75\\72.16 $\pm$ 0.86\hspace{0.4cm} 75.94 $\pm$ 0.76\\70.89 $\pm$ 0.67\hspace{0.4cm} 72.54 $\pm$ 0.77\\\textbf{73.45 $\pm$ 0.93}\hspace{0.15cm} \textbf{76.80 $\pm$ 0.81}}&\makecell[l] {79.11 $\pm$ 0.70\hspace{0.4cm} 75.59 $\pm$ 0.65\\83.96 $\pm$ 0.64\hspace{0.4cm} 82.88 $\pm$ 0.67\\\textbf{85.46 $\pm$ 0.47}\hspace{0.15cm} 75.97 $\pm$ 0.85\\81.65 $\pm$ 0.64\hspace{0.4cm} 81.31 $\pm$ 0.66\\81.33 $\pm$ 0.68\hspace{0.4cm} 81.46 $\pm$ 0.65\\85.33 $\pm$ 0.56\hspace{0.4cm} 81.96 $\pm$ 0.70\\82.84 $\pm$ 0.66\hspace{0.27cm} \textbf{82.96 $\pm$ 0.62}} \\
         \hline
    \end{tabular}}
    \label{tab:fsor_c}
\end{table*}

\noindent\textit{Evaluation metrics.} Following existing FSOR methods~\citep{liu2020few, wang2023glocal}, we use ACC and AUROC as the metrics to measure the effectiveness of our model. ACC is used to measure the classification accuracy of the closed-set samples. It can be calculated by dividing the number of correctly classified samples in $\mathcal{Q}_s$ by the number of samples in $\mathcal{Q}_s$. AUROC is used to measure the accuracy of open-set recognition. It is the area under the ROC curve for open-set recognition of all query samples $\mathcal{Q}_s \cup \mathcal{Q}_u$. A larger ACC or AUROC indicates better performance.

\noindent\textit{Implementation details.}
Following previous FSOR methods~\citep{deng2022learning,huang2022task, jeong2021few}, we use ResNet-12~\citep{he2016deep} as the feature extractor of our network. We first use FEAT to pre-train the backbone on the training set, then
remove the last fully connected layer and the softmax layer. The remaining layers are fixed and used as feature extractors in our meta-learning process. Similar to class-incremental learning, extracted features are considered to be located in the tangent space at the origin and then projected to the hyperbolic space via an exponential map added on the top of the backbone. For an N-way K-shot FSOR task, we follow previous FSOR methods~\citep{deng2022learning,huang2022task, jeong2021few} to select 2N classes in the dataset, half of which are seen classes and the other are unseen classes. For seen classes, we sample K images for each class as a support set and another 15 images as a seen query set. For each unseen class, we only
sample 15 images as an unseen query set. During the training phase, we perform our hyperbolic dual feature augmentation using the seen classes to synthesize novel features of unseen classes to train the feature extractor. During the testing phase, the first 75 query samples with the highest open-set scores are taken as open-set samples to calculate the open-set recognition metrics.

\noindent\textit{Hyperparameters.} 
In the meta-learning stage, we use the Adam optimizer to train the model for over 60,000 tasks. The learning rate is set to 1e-6 for feature extractor and 1e-4 for our hyperbolic dual feature augmentation module (\emph{i.e.}, gradient flow networks $F_{1}$, $F_{2}$, $F_{3}$, $F_{4}$, $F_{5}$ and $F_{6}$) and decayed by the factor of 10 for every 12,000
tasks. The trade-off parameter $\gamma$ in the hierarchy-preserving regularization loss is set as 0.1. The initial value of the
curvature c is set as -0.1. We then select the model with the best results through the validation set for testing.

\begin{table*}[t]
    \centering

    \caption{Accuracy (\%) comparisons with augmentation-based few-shot learning methods on the mini-ImageNet, CUB, and CIFAR-FS datasets. ‘Euclidean’ and ‘Hyperbolic’ denote the algorithms are in Euclidean spaces and hyperbolic spaces, respectively.}
    \renewcommand{\arraystretch}{1.3}
    \scalebox{0.67}{\begin{tabular}{l|ccccc}
        \hline
         Methods & Backbone &Category &\makecell[c]{mini-ImageNet \\ 1-shot\hspace{0.8cm} 5-shot} & \makecell[c]{CUB \\ 1-shot\hspace{0.8cm} 5-shot}& \makecell[c]{CIFAR-FS \\ 1-shot\hspace{0.8cm} 5-shot} \\
         \hline
         VFSL~\citep{zhang2019variational} & ResNet12&Euclidean&
         \makecell[c]{61.23$\pm$0.26\hspace{0.3cm}77.69$\pm$0.17}&
         \makecell[c]{-\hspace{1.8cm}-}&
         \makecell[c]{-\hspace{1.8cm}-}\\
         DTN~\citep{chen2020diversity} & ResNet12&Euclidean&
         \makecell[c]{63.45$\pm$0.86\hspace{0.3cm}77.91$\pm$0.62}&
         \makecell[c]{72.00\hspace{1.2cm}85.10}&
         \makecell[c]{71.50\hspace{1.2cm}82.80}\\
         Meta-HP~\citep{zhang2023meta} & ResNet12&Euclidean&
         \makecell[c]{65.08$\pm$0.73\hspace{0.3cm}79.40$\pm$0.39}&
         \makecell[c]{-\hspace{1.8cm}-}&
         \makecell[c]{73.74$\pm$0.57\hspace{0.3cm}86.37$\pm$0.32}\\
         HFA~\citep{gao2022hyperbolic} & ResNet12&Hyperbolic&
         \makecell[c]{66.87$\pm$0.44\hspace{0.3cm}82.08$\pm$0.31}&
         \makecell[c]{76.75$\pm$0.43\hspace{0.3cm}89.19$\pm$0.46}&
         \makecell[c]{72.52$\pm$0.46\hspace{0.3cm}85.33$\pm$0.33}\\
         \textbf{HDFA (Ours)} & ResNet12&Hyperbolic&
         \makecell[c]{\textbf{67.12$\pm$0.45}\hspace{0.3cm}\textbf{83.04$\pm$0.42}}&
         \makecell[c]{\textbf{78.23$\pm$0.55}\hspace{0.3cm}\textbf{91.50$\pm$0.31}}&
         \makecell[c]{\textbf{73.90$\pm$0.81}\hspace{0.3cm}\textbf{86.60$\pm$0.52}}\\
         \hline
         $\Delta$-encoder~\citep{schwartz2018delta} & VGG16 &Euclidean&
         \makecell[c]{59.90\hspace{1.2cm}69.70}&
         \makecell[c]{69.80$\pm$0.46\hspace{0.3cm}82.60$\pm$0.35}&
         \makecell[c]{66.70\hspace{1.2cm}79.80}\\
         SalNet~\citep{zhang2019few} & ResNet101&Euclidean&
         \makecell[c]{62.22$\pm$0.87\hspace{0.3cm}77.95$\pm$0.65}&
         \makecell[c]{-\hspace{1.8cm}-}&
         \makecell[c]{-\hspace{1.8cm}-}\\
         Dual TriNet~\citep{chen2019multi} & ResNet18&Euclidean&
         \makecell[c]{58.12$\pm$1.37\hspace{0.3cm}82.32$\pm$.0.45}&
         \makecell[c]{69.61\hspace{1.2cm}84.10}&
         \makecell[c]{63.41$\pm$0.64\hspace{0.3cm}78.43$\pm$0.64}\\
         IDeMe-Net~\citep{chen2019image} & ResNet18&Euclidean&
         \makecell[c]{59.14$\pm$0.86\hspace{0.3cm}74.63$\pm$0.74}&
         \makecell[c]{-\hspace{1.8cm}-}&
         \makecell[c]{-\hspace{1.8cm}-}\\
         V1-Net~\citep{Luo2021FewShotLV} & ResNet18&Euclidean&
         \makecell[c]{61.05\hspace{1.2cm}78.60}&
         \makecell[c]{74.76\hspace{1.2cm}86.84}&
         \makecell[c]{-\hspace{1.8cm}-}\\
         AFHN~\citep{Li2020AdversarialFH} & ResNet18&Euclidean&
         \makecell[c]{62.38$\pm$0.72\hspace{0.3cm}78.16$\pm$0.56}&
         \makecell[c]{70.53$\pm$1.01\hspace{0.3cm}83.95$\pm$0.63}&
         \makecell[c]{68.32$\pm$0.93\hspace{0.3cm}81.45$\pm$0.87}\\
         TFH~\citep{Lazarou2021TensorFH}& ResNet18&Euclidean&
         \makecell[c]{65.07$\pm$0.82\hspace{0.3cm}80.81$\pm$0.61}&
         \makecell[c]{75.76$\pm$0.83\hspace{0.3cm}88.60$\pm$0.74}&
         \makecell[c]{74.77$\pm$0.90\hspace{0.3cm}86.88$\pm$0.59}\\
         \textbf{HDFA (Ours)} & ResNet18&Hyperbolic&
         \makecell[c]{\textbf{68.61$\pm$0.49}\hspace{0.3cm}\textbf{83.74$\pm$0.32}}&
         \makecell[c]{\textbf{78.79$\pm$0.46}\hspace{0.3cm}\textbf{92.04$\pm$0.28}}&
         \makecell[c]{\textbf{75.37$\pm$0.50}\hspace{0.3cm}\textbf{87.92$\pm$0.57}}\\
         \hline
    \end{tabular}}
    \label{tab:few_aug}
\end{table*}

\begin{table}[t]
    \centering
    \caption{Accuracy (\%) comparisons with popular data augmentation methods on the FC100 dataset. All methods use ResNet12 as the backbone. }
    \begin{tabular}{lc}
        \hline
         Methods & \makecell[c]{FC100 \\ 1-shot\hspace{0.5cm} 5-shot}  \\
         \toprule
         Mixup~\citep{thulasidasan2019mixup}&
         \makecell[c]{41.30\hspace{0.7cm}55.32}\\
         \specialrule{0em} {1.5pt}{1.5pt}
         MetaMix~\citep{yao2021improving}&
         \makecell[c]{41.42\hspace{0.7cm}54.93}\\
         \specialrule{0em} {1.5pt}{1.5pt}
         MLTI~\citep{yao2021meta}&
         \makecell[c]{42.01\hspace{0.7cm}55.41}\\
         \specialrule{0em} {1.5pt}{1.5pt}
         classAug-Rotation~\citep{zhu2022learning}&
         \makecell[c]{42.37\hspace{0.7cm}56.55}\\
         \specialrule{0em} {1.5pt}{1.5pt}
         classAug-BCI~\citep{zhu2022learning}&
         \makecell[c]{43.02\hspace{0.7cm}57.25}\\
         \specialrule{0em} {1.5pt}{1.5pt}
         HFA~\citep{gao2022hyperbolic}&
         \makecell[c]{44.04\hspace{0.7cm}58.11}\\
         \specialrule{0em} {1.5pt}{1.5pt}
         \textbf{HDFA (Ours)} &
         \makecell[c]{\textbf{45.17}\hspace{0.7cm}\textbf{59.34}}\\
         \hline
    \end{tabular}
    \label{tab:few_FC100}
\end{table}

\subsubsection{Results}

Table~\ref{tab:fsor} and Table~\ref{tab:fsor_c} show a comparison of our method, HDFA, with other FSOR methods on the mini-ImageNet and CIFAR-FS datasets under both 1-shot and 5-shot settings. The results demonstrate that our method significantly outperforms previous FSOR methods. Specifically, HDFA achieves the best open-set recognition performance, measured by AUROC, across both datasets in both settings, while simultaneously maintaining highly competitive closed-set classification accuracy.
For instance, on the mini-ImageNet dataset, HDFA improves the AUROC by 3.85\% and 2.83\% compared to the state-of-the-art methods OPP~\citep{sun2024overall} and GEL~\citep{wang2023glocal} in the 1-shot task, and by 2.83\% and 2.30\% in the 5-shot task, respectively. Additionally, when compared to HFA~\citep{gao2022hyperbolic}, which performs feature augmentation only for seen classes in the hyperbolic space, our method not only surpasses HFA in closed-set classification but also achieves better open-set recognition results.
For example, on the CIFAR-FS dataset, HDFA improves the closed-set classification accuracy (ACC) by 1.29\% and 1.51\%, and the open-set recognition accuracy (AUROC) by 0.86\% and 1.50\% in the 1-shot and 5-shot tasks, respectively. These results indicate the advantage of incorporating additional unseen classes during training, which improves the learned
feature representation and enhances the open-environment generalization.

\subsection{Few-Shot Learning}
\subsubsection{Task Setting}
In few-shot learning (FSL), given a labeled dataset of base classes with a large number of images, the goal is to learn concepts in novel classes with a few samples. In an $N$-way $K$-shot few-shot learning task, the support set contains $N$ classes with $K$ samples per class, the query set contains samples from the same $N$ classes with $Q$ samples per class, and the goal is to classify the $N \times Q$ query images into $N$ classes.
\subsubsection{Experimental Setup}
\textit{Datasets.} We conduct experiments on Five few-shot learning datasets: mini-ImageNet~\citep{vinyals2016matching}, tiered-ImageNet~\citep{ren2018meta}, CUB~\citep{wah2011caltech}, CIFAR-FS~\citep{krizhevsky2009learning} and FC100~\citep{liu2020prototype} datasets. 
The tiered-ImageNet dataset has 779165
images from 608 classes, involving 351 training classes, 97 validation classes, and 160 test classes. The CUB dataset is a fine-grained image dataset that contains 11788 images from 200 bird
classes, and is split into 100, 50, and 50 classes
for training, validation, and testing, respectively. 
We crop bird regions from images with the given
bounding boxes before training. 
The FC100 dataset is a split dataset based on CIFAR-100~\citep{krizhevsky2009learning}, containing 60 base classes for training, 20 classes for validation, and 20 novel classes for testing. Each class consists of 600 images. These five datasets show relatively obvious hyperbolic structures and are commonly used to evaluate hyperbolic algorithms~\citep{gao2021curvature,moreira2024hyperbolic,yue2024understanding}.

\noindent\textit{Baselines.} 
We compare HDFA with the following augmentation-based few-shot learning methods: Dual TriNet~\citep{chen2019multi}, IDeMe-Net~\citep{chen2019image}, V1-Net~\citep{Luo2021FewShotLV}, AFHN~\citep{Li2020AdversarialFH}, TFH~\citep{Lazarou2021TensorFH}, Delta-encoder~\citep{schwartz2018delta}, SalNet~\citep{zhang2019few}, VFSL~\citep{zhang2019variational}, DTN~\citep{chen2020diversity} and Meta-HP~\citep{zhang2023meta} on the mini-ImageNet, CUB and CIFAR-FS datasets. We compare HDFA with other popular augmentation
methods such as mixup~\citep{thulasidasan2019mixup}, MetaMix~\citep{yao2021improving}, MLTI~\citep{yao2021meta} and classAug~\citep{zhu2022learning} on the FC100 dataset. We also compare our method with popular optimization-based few-shot learning methods: MAML~\citep{finn2017model}, L2F~\citep{baik2020learning},
ALFA~\citep{baik2020meta}, MeTAL~\citep{baik2021meta}, and CurAML~\citep{gao2022curvature}, and metric-based few-shot learning methods: ProtoNet~\citep{snell2017prototypical}, HyperProto~\citep{khrulkov2020hyperbolic}, MetaOptNet~\citep{lee2019meta}, DSN~\citep{simon2020adaptive}, FEAT~\citep{ye2020few}, CGCS~\citep{gao2021curvature} and DiffKendall~\citep{zheng2023diffkendall} on the mini-ImageNet and tiered-ImageNet datasets.

\noindent\textit{Implementation details.}
In our experiments, we employ the pipeline in the hyperbolic prototypical network~\citep{khrulkov2020hyperbolic} and perform augmentation for the support data. Specifically, we use ResNet12~\citep{he2016deep} and ResNet18~\citep{he2016deep} as the backbone networks. We first pre-train the two backbones on the training set, then remove the last fully connected layer and the softmax layer. The remaining layers are fixed and used as feature extractors in our meta-learning process. Similar to the class incremental and few-shot open-set tasks, we use an exponential
map to project the features extracted from the feature extractor to the hyperbolic space. We follow the standard N-way K-shot (\emph{e.g.}, 5-way 1-shot and 5-way 5-shot) protocol processed by~\citep{vinyals2016matching} to evaluate the model.

\noindent\textit{Hyperparameters.} In the 1-shot setting, the ratio of the training set $\mathcal{D}_t$ to the validation set $\mathcal{D}_v$ is 1:15, where we sample one sample per class for training and 15 samples for validation. In the 5-shot setting, the ratio of $\mathcal{D}_t$ to $\mathcal{D}_v$ is 1:3, where we sample 5 samples per class for training and 15 samples for validation. The initial value of the curvature $c$ is set as -0.3.

\begin{table*}[t]
    \centering

    \caption{Accuracy (\%) comparisons with popular few-shot learning methods on the mini-ImageNet and tiered-ImageNet datasets. ‘Optim’, ‘Metric’, and ‘Aug’ denote the optimization-based, metric-based, and augmentation-based few-shot learning methods, respectively. ‘Euclidean’ and ‘Hyperbolic’ denote the algorithms are in Euclidean spaces and hyperbolic spaces, respectively. $*$ means that HyperProto uses ResNet18 as the backbone, while the others use ResNet12.}
    \textbf{\renewcommand{\arraystretch}{1.5}}
    \scalebox{0.87}{\begin{tabular}{l|ccc}
        \hline
         Methods & Category & \makecell[c]{mini-ImageNet \\ 1-shot\hspace{0.8cm} 5-shot} & \makecell[c]{tiered-ImageNet \\ 1-shot\hspace{0.8cm} 5-shot} \\
         \hline
         MAML~\citep{finn2017model} &Euclidean Optim&
         \makecell[c]{51.03$\pm$0.50\hspace{0.3cm}68.26$\pm$0.47}&         \makecell[c]{58.58$\pm$0.49\hspace{0.3cm}71.24$\pm$0.43}\\
         L2F~\citep{baik2020learning} &Euclidean Optim&
         \makecell[c]{57.48$\pm$0.49\hspace{0.3cm}74.68$\pm$0.43}&
         \makecell[c]{63.94$\pm$0.84\hspace{0.3cm}77.61$\pm$0.41}\\
         ALFA~\citep{baik2020meta} &Euclidean Optim&
         \makecell[c]{60.06$\pm$0.49\hspace{0.3cm}74.68$\pm$0.43}&
         \makecell[c]{64.43$\pm$0.49\hspace{0.3cm}81.77$\pm$0.39}\\
         MeTAL~\citep{baik2021meta} &Euclidean Optim&
         \makecell[c]{59.64$\pm$0.38\hspace{0.3cm}76.20$\pm$0.19}&
         \makecell[c]{63.89$\pm$0.43\hspace{0.3cm}80.14$\pm$0.40}\\
         CurAML~\citep{gao2022curvature} &Hyperbolic Optim&
         \makecell[c]{63.30$\pm$0.41\hspace{0.3cm}81.04$\pm$0.39}&
         \makecell[c]{68.46$\pm$0.56\hspace{0.3cm}83.84$\pm$0.40}\\
         \hline
         ProtoNet~\citep{snell2017prototypical} &Euclidean Metric&
         \makecell[c]{56.52$\pm$0.45\hspace{0.3cm}74.28$\pm$0.20}&
         \makecell[c]{53.51$\pm$0.89\hspace{0.3cm}72.69$\pm$0.74}\\
         MetaOptNet~\citep{lee2019meta} & Euclidean Metric&
         \makecell[c]{62.64$\pm$0.61\hspace{0.3cm}78.63$\pm$0.46}&
         \makecell[c]{65.99$\pm$0.72\hspace{0.3cm}81.56$\pm$0.53}\\
         DSN~\citep{simon2020adaptive} & Euclidean Metric&
         \makecell[c]{62.64$\pm$0.66\hspace{0.3cm}78.83$\pm$0.45}&
         \makecell[c]{66.22$\pm$0.75\hspace{0.3cm}82.79$\pm$0.48}\\
         FEAT~\citep{ye2020few} &Euclidean Metric&
         \makecell[c]{66.78$\pm$0.20\hspace{0.3cm}82.05$\pm$0.14}&
         \makecell[c]{70.80$\pm$0.23\hspace{0.3cm}84.79$\pm$0.16}\\
         HyperProto$*$~\citep{khrulkov2020hyperbolic} & Hyperbolic Metric&
         \makecell[c]{59.74$\pm$0.20\hspace{0.3cm}76.84$\pm$0.14}&
         \makecell[c]{-\hspace{1.8cm}-}\\
         CGCS~\citep{gao2021curvature} & Hyperbolic Metric&
         \makecell[c]{67.02$\pm$0.20\hspace{0.3cm}82.32$\pm$0.14}&
         \makecell[c]{71.66$\pm$0.23\hspace{0.3cm}85.50$\pm$0.15}\\
         DiffKendall~\citep{zheng2023diffkendall} &Euclidean Metric&
         \makecell[c]{65.56$\pm$0.43\hspace{0.3cm}80.79$\pm$0.31}&
         \makecell[c]{70.76$\pm$0.43\hspace{0.3cm}85.31$\pm$0.34}\\
         
         \hline
         HFA~\citep{gao2022hyperbolic} & Hyperbolic Aug&
         \makecell[c]{66.87$\pm$0.44\hspace{0.3cm}82.08$\pm$0.31}&
         \makecell[c]{71.62$\pm$0.49\hspace{0.3cm}85.47$\pm$0.35}\\
         \textbf{HDFA (Ours)} & Hyperbolic Aug&
         \makecell[c]{\textbf{67.12$\pm$0.45}\hspace{0.3cm}\textbf{83.04$\pm$0.42}}&
         \makecell[c]{\textbf{71.94$\pm$0.47}\hspace{0.3cm}\textbf{85.88$\pm$0.34}}\\
         
         \hline
    \end{tabular}}
    \label{tab:few_meta}
\end{table*}

In the pre-training stage, we train the two backbones for 120 epochs using the SGD optimizer. The learning rate is set as 0.01, decaying by 0.1 every 40 epochs. Then, we remove the last fully connected layer and the softmax layer from the pre-trained model, using the fixed remaining layers as our feature extractor. In the meta-learning stage, we train our hyperbolic dual feature augmentation module (\emph{i.e.}, gradient flow networks $F_{1}$, $F_{2}$, $F_{3}$, $F_{4}$, $F_{5}$ and $F_{6}$) for 4000 episodes with the Adam optimizer, using a learning rate of 0.001 that decays by 0.1 every 2000 episodes. The weight decay for the feature extractor is set as 0.0005. The trade-off parameter $\gamma$ in the hierarchy-preserving regularization loss is set as 0.1.

\subsubsection{Results}
Table~\ref{tab:few_aug} shows the accuracy comparisons of our method and other augmentation-based few-shot learning methods on the mini-ImageNet, CUB, and CIFAR-FS datasets. Unlike our method performing feature augmentation in the hyperbolic space, VFSL~\citep{zhang2019variational} and DTN~\citep{chen2020diversity} perform feature augmentation and other methods perform data augmentation all in the Euclidean space. As shown in the table, our method brings improvements on them. For example, compared with VFSL using ResNet12 on the mini-ImageNet dataset, our method improves the accuracies by 5.89\% and 5.35\% respectively in the 1-shot and 5-shot tasks. This suggests that augmentation in the hyperbolic
space better explores and captures the latent hierarchical structures of the scarce data and obtains a more accurate estimation of the true distribution. Results on ResNet18 also demonstrate the superiority of our method. Particularly, the proposed HDFA achieves better performance than feature augmentation only for seen classes in the hyperbolic space (\emph{i.e.}, HFA). For example, on the CUB dataset using ResNet18, our method improves HFA by 1.54\% and 1.27\% in the 1-shot and 5-shot tasks, respectively. This indicates that introducing additional unseen classes can improve the learned feature representation, which is beneficial for new class generalization in few-shot
learning.

Table~\ref{tab:few_FC100} shows the accuracy comparisons of our method and some popular data augmentation methods on the FC100 dataset. The classAug method and our HDFA introduce unseen classes, while other augmentation methods only generate samples for original seen classes. Our method performs feature augmentation in the hyperbolic space, while others all perform data augmentation in the Euclidean space. Compared with traditional data augmentation methods, such as Mixup~\citep{thulasidasan2019mixup} and MetaMix~\citep{yao2021meta}, our method remarkably shows stronger performance. For example, our method improves MetaMix by 3.00\% and 4.03\% in the 1-shot and 5-shot tasks, respectively. As demonstrated in the table, our method also shows superior performance compared to classAug. The main reason is that classAug simply uses the rotation and linear between-class interpolation strategies to generate samples for unseen classes, while our method learns neural networks to estimate appropriate distributions for unseen classes with respect to the underlying geometric structure of the data. In this case, our method can better adapt to complex distributions in scarce data regimes and avoid undesirable data distortion.

We then compare HDFA with popular optimization-based and metric-based few-shot learning methods
on the mini-ImageNet and tiered-ImageNet datasets. Results are shown in Table~\ref{tab:few_meta}. We observe that our performance achieves better performance than these methods. For example, we achieve 67.12\% and 83.04\% on the 1
shot and 5-shot tasks on the mini-ImageNet dataset, 1.56\% and 2.25\% higher than DiffKendall~\citep{zheng2023diffkendall}. When compared with CurAML~\citep{gao2022curvature}, HyperProto~\citep{khrulkov2020hyperbolic}, CGCS~\citep{gao2021curvature}
that are all few-shot learning methods in the hyperbolic space, we also achieve satisfying performance. Practically, our method improves the plain HyperProto method by 7.38\% and 6.20\%, which indicates that
our dual data augmentation can enhance the model in scarce data regimes.

\subsection{Zero-Shot Learning}
\subsubsection{Task Setting}
Zero-shot learning (ZSL) aims to recognize classes that have no labeled data in the training set. $\mathcal{X}^s$ and $\mathcal{X}^u$ denote the set of image visual features, $\mathcal{Y}^s$ and $\mathcal{Y}^u$ the set of image class labels, for seen $(s)$ and unseen $(u)$ classes, respectively. The training set for seen classes is represented as $\mathcal{D}^s=\left\{x_i^s, y_i^s\right\}_{i=1}^{N^s}$ of $N^s$ elements and has $M^s$ classes in total. The test set for unseen classes is similarly represented as $\mathcal{D}^u=\left\{x_i^u, y_i^u\right\}_{i=1}^{N^u}$ for $N^u$ elements and has $M^u$ classes in total. There are in total $K$ object attributes shared between seen and unseen classes while there is no overlap between $\mathcal{D}^s$ and $\mathcal{D}^u$. In the conventional ZSL, the task is to recognize only unseen classes. In the more realistic generalized zero-shot learning (GZSL), the task is to recognize both the seen and unseen classes. 

\subsubsection{Experimental Setup}
\textit{Dataset and evaluation metrics.}
We evaluate our method on the CUB~\citep{wah2011caltech} dataset. We report results in both ZSL and GZSL settings. In the ZSL setting, we only evaluate the performance on unseen classes and use the top-1 accuracy ($T_{1}$) as the evaluation metric. In the GZSL setting, we evaluate the performance on both seen and unseen classes and follow~\citep{xian2018zero} to use generalized seen accuracy ($Acc_{S}$), generalized unseen accuracy ($Acc_{U}$) and their generalized Harmolic Mean ($Acc_{H}$) as evaluate metrics. The former two are top-1 accuracy for seen and unseen classes respectively while the last one is obtained by,
\begin{equation}
A c c_H=\frac{2 \cdot A c c_U \cdot A c c_S}{A c c_U+A c c_S} .
\tag{26}
\end{equation}
$Acc_{H}$ measures the inherent bias towards seen classes, which is a more important metric in GZSL.

\noindent\textit{Baselines.} 
We compare our method with the following generative-based methods: OCD-CVAE~\citep{keshari2020generalized}, LsrGAN~\citep{vyas2020leveraging}, FREE~\citep{chen2021free}, E-PGN~\citep{yu2020episode}, GCM-CF~\citep{yue2021counterfactual}, TransferIF~\citep{feng2020transfer}, HSVA~\citep{chen2021hsva} and ICCE~\citep{kong2022compactness}. We also compare our method with non-generative-based methods: CN-ZSL~\citep{skorokhodov2020class}, APZ~\citep{zhang2019probabilistic}, DVBE~\citep{min2020domain}, AREN~\citep{xie2019attentive}, RGEN~\citep{xie2020region}, DAZLE~\citep{huynh2020fine}, and APN~\citep{xu2020attribute}. We compare HDFA with the above methods in both ZSL and GZSL settings.

\begin{table*}[t]
    \centering

    \caption{Accuracy (\%) comparisons with zero-shot learning methods on the CUB dataset. We report Top-1 accuracy $T_{1}$ for ZSL, $Acc_U$, $Acc_S$, $Acc_H$ for GZSL. ‘Euclidean’ and ‘Hyperbolic’ denote the algorithms are in Euclidean spaces and hyperbolic spaces, respectively.}
    \renewcommand{\arraystretch}{1.3}
    \scalebox{0.85}{\begin{tabular}{l|lcc}
        \hline
         &Methods &Category &\makecell[c]{CUB\\$T_{1}$\hspace{1cm}$Acc_U$\hspace{0.5cm} $Acc_S$\hspace{0.5cm} $Acc_H$ } \\
         \hline
        \makecell[l]{Non-generative}&\makecell[l]{CN-ZSL~\citep{skorokhodov2020class}\\ APZ~\citep{zhang2019probabilistic}\\ DVBE~\citep{min2020domain}\\ AREN~\citep{xie2019attentive}\\   RGEN~\citep{xie2020region}\\ DAZLE~\citep{huynh2020fine}\\ APN~\citep{xu2020attribute}}&\makecell[c]{Euclidean\\Euclidean\\Euclidean\\Euclidean\\Euclidean\\Euclidean\\Euclidean}&\makecell[l]{-\hspace{1.69cm} 49.90\hspace{0.5cm} 50.70 \hspace{0.5cm}50.30\\53.20\hspace{1cm} 58.20 \hspace{0.5cm}37.80\hspace{0.5cm} 45.90\\-\hspace{1.69cm} 53.20\hspace{0.5cm} 60.20\hspace{0.5cm} 56.50\\ 71.80\hspace{1cm} 63.20\hspace{0.5cm} 69.00\hspace{0.5cm} 66.00\\76.10\hspace{1cm} 60.00\hspace{0.5cm} 73.50\hspace{0.5cm} 66.10\\ 66.00\hspace{1cm} 56.70\hspace{0.5cm} 59.60\hspace{0.5cm} 58.10 \\72.00\hspace{1cm} 65.30\hspace{0.5cm} 69.30\hspace{0.5cm} 67.20} \\
         \hline
         \makecell[l]{Generative}&\makecell[l]{OCD-CVAE~\citep{keshari2020generalized}\\              LsrGAN~\citep{vyas2020leveraging}\\ FREE~\citep{chen2021free}\\E-PGN~\citep{yu2020episode}\\ GCM-CF~\citep{yue2021counterfactual}\\ TransferIF~\citep{feng2020transfer}\\ HSVA~\citep{chen2021hsva}\\ ICCE~\citep{kong2022compactness}\\HFA~\citep{gao2022hyperbolic}\\\textbf{HDFA (Ours)}}&\makecell[c]{Euclidean\\Euclidean\\Euclidean\\Euclidean\\Euclidean\\Euclidean\\Euclidean\\Euclidean\\Hyperbolic\\Hyperbolic}&\makecell[l]{ 60.30\hspace{1cm} 44.80\hspace{0.5cm} 59.90\hspace{0.5cm} 51.30\\ 60.30\hspace{1cm} 48.10\hspace{0.5cm} 59.10\hspace{0.5cm} 53.00\\-\hspace{1.69cm} 55.70\hspace{0.5cm} 59.90\hspace{0.5cm} 57.70\\ 72.40\hspace{1cm} 52.00\hspace{0.5cm} 61.10\hspace{0.5cm} 56.20\\-\hspace{1.69cm} 61.00\hspace{0.5cm} 59.70\hspace{0.5cm} 60.30\\-\hspace{1.69cm} 52.10\hspace{0.5cm} 53.30\hspace{0.5cm} 52.70\\-\hspace{1.69cm} 52.70\hspace{0.5cm} 58.30\hspace{0.5cm} 55.30\\-\hspace{1.69cm} 67.30\hspace{0.5cm} 65.50\hspace{0.5cm} 66.40\\ 76.17\hspace{1cm} 67.54\hspace{0.5cm} 75.46\hspace{0.5cm} 71.28\\ \textbf{76.56}\hspace{0.8cm} \textbf{68.09}\hspace{0.4cm} \textbf{76.56}\hspace{0.4cm} \textbf{72.07}}\\
         \hline
    \end{tabular}}
    \label{tab:zero}
\end{table*}
\begin{table}[t]
    \centering
    \caption{Accuracy (\%) results on the general image classification experiment using the CIFAR10 dataset. }
    \begin{tabular}{lcc}
        \hline
         Methods & Backbone &\makecell[c]{CIFAR10}  \\
         \toprule
         \specialrule{0em} {1.5pt}{1.5pt}
         Hyperbolic Baseline& WDN-28-10& 
         \makecell[c]{92.92}\\
         \specialrule{0em} {1.5pt}{1.5pt}
         HFA~\citep{gao2022hyperbolic}& WDN-28-10& 
         \makecell[c]{93.17}\\
         \specialrule{0em} {1.5pt}{1.5pt}
         \textbf{HDFA (Ours)}& WDN-28-10& 
         \makecell[c]{\textbf{93.45}}\\
         \hline
    \end{tabular}
    \label{tab:gen}
\end{table}

\noindent\textit{Implementation details.} In our experiments, we choose the ResNet101~\citep{he2016deep} as the backbone network. The input image is randomly cropped on a 224 $\times$ 224 resized image with random horizontal flipping. We first pre-train the backbone on the ImageNet1k dataset~\citep{deng2009imagenet} (1.28 million images, 1000 classes), then remove the last fully connected layer and the softmax layer. The remaining layers are fixed and used as feature extractors in our meta-learning process. Similar to the above few-shot learning task, we use an exponential map to project the features extracted from the feature extractor to the hyperbolic space. 

\noindent\textit{Hyperparameters.} We follow the proposed train and test split in~\citep{xian2018zero}. The initial value of the curvature $c$ is set as -0.001. We choose the Adam optimizer to train the model. The learning rate is set to 1e-5 for the feature extractor and 1e-3 for our hyperbolic dual feature augmentation module (i.e., gradient flow networks F1, F2, F3, F4, F5, and F6). The weight decay is set to 0.0005. The trade-off parameter $\gamma$ in the hierarchy-preserving regularization loss is set as 0.1. 

\begin{table*}[t]
    \centering

    \caption{Ablation study of the proposed method on the Tiny-ImageNet dataset with different numbers of incremental stages. All methods use ResNet18 as the backbone.}
    \scalebox{0.9}{ \begin{tabular}{lc}
        \hline
         Methods&\makecell[c]{Tiny-ImageNet \\ T200-B100-S5\hspace{0.4cm} T200-B100-S10 \hspace{0.4cm} T200-B100-S20}  \\
         \toprule
         \specialrule{0em} {1.5pt}{1.5pt}
         w/o Aug &
         \makecell[c]{34.80\hspace{2cm}31.18\hspace{2cm}30.96}\\
         \specialrule{0em} {1.5pt}{1.5pt}
         Inf Aug &
         \makecell[c]{35.58\hspace{2cm}32.24\hspace{2cm}32.22}\\
         \specialrule{0em} {1.5pt}{1.5pt}
         Neural ODE (seen)+Aug 5 samples &
         \makecell[c]{35.82\hspace{2cm}32.63\hspace{2cm}32.90}\\
         \specialrule{0em} {1.5pt}{1.5pt}
         HFA (seen+Neural ODE+Inf Aug)  &\makecell[c]{36.11\hspace{2cm}33.65\hspace{2cm}33.73}\\
     	\specialrule{0em} {1.5pt}{1.5pt}
         HDFA (seen+unseen+Neural ODE+Inf Aug) & \makecell[c]{\textbf{38.23}\hspace{1.9cm}\textbf{36.34}\hspace{1.85cm}\textbf{35.15}}\\
         \hline
    \end{tabular}}
    \label{tab:final_tiny_abla}
\end{table*}

\begin{table*}[t]
    \centering
    \caption{Ablation study of the proposed method on the mini-ImageNet and tiered-ImageNet dataset. All methods use ResNet12 as the backbone. }
    \scalebox{0.88}{\begin{tabular}{lcc}
        \hline
         Methods  &\makecell[c]{mini-ImageNet \\ 1-shot\hspace{0.5cm} 5-shot} & \makecell[c]{tiered-ImageNet \\ 1-shot\hspace{0.5cm} 5-shot}  \\
         \toprule
         \specialrule{0em} {1.5pt}{1.5pt}
         w/o Aug& 
         \makecell[c]{64.75$\pm$0.35\hspace{0.3cm}79.84$\pm$0.20}&
         \makecell[c]{68.57$\pm$0.49\hspace{0.3cm}82.22$\pm$0.39}\\
         \specialrule{0em} {1.5pt}{1.5pt}
         Inf Aug& 
         \makecell[c]{65.12$\pm$0.45\hspace{0.3cm}80.78$\pm$0.32}&
         \makecell[c]{70.22$\pm$0.50\hspace{0.3cm}84.51$\pm$0.37}\\
         \specialrule{0em} {1.5pt}{1.5pt}
         Neural ODE (seen)+Aug 5 samples& 
         \makecell[c]{65.54$\pm$0.46\hspace{0.3cm}80.96$\pm$0.31}&
         \makecell[c]{69.32$\pm$0.53\hspace{0.3cm}83.21$\pm$0.35}\\
         \specialrule{0em} {1.5pt}{1.5pt}
         HFA (seen+Neural ODE+Inf Aug) & 
         \makecell[c]{66.87$\pm$0.44\hspace{0.3cm}82.08$\pm$0.31}&
         \makecell[c]{71.62$\pm$0.49\hspace{0.3cm}85.47$\pm$0.35}\\
         \specialrule{0em} {1.5pt}{1.5pt}
         HDFA (seen+unseen+Neural ODE+Inf Aug) & 
         \makecell[c]{\textbf{67.12$\pm$0.45}\hspace{0.3cm}\textbf{83.04$\pm$0.42}}&
         \makecell[c]{\textbf{71.94$\pm$0.47}\hspace{0.3cm}\textbf{85.88$\pm$0.34}}\\
         \hline
    \end{tabular}}
    \label{tab:abaltion}
\end{table*}

\subsubsection{Results}
Table~\ref{tab:zero} presents a comparison of the proposed method HDFA, with other zero-shot learning (ZSL) methods on the CUB dataset. The results indicate that HDFA consistently outperforms both generative and non-generative ZSL methods across most evaluation metrics. Notably, for top-1 accuracy ($T_{1}$) and the generalized harmonic mean ($Acc_{H}$), which are critical indicators for ZSL and GZSL, HDFA achieves 76.56\% for $T_{1}$ and 72.07\% for $Acc_{H}$, respectively. These results demonstrate the effectiveness of our method in tackling the challenges of zero-shot learning in open-environment settings. Additionally, when compared to HFA~\citep{gao2022hyperbolic}, which performs feature augmentation exclusively for seen classes in hyperbolic space, our HDFA improves $T_{1}$ by 0.39\% for ZSL and achieves notable improvements for GZSL, including 0.55\% for $Acc_{S}$, 1.10\% for $Acc_{U}$, and 0.79\% for $Acc_{H}$. This indicates the advantage of introducing features of unseen classes during training, as it enhances the model’s ability to generalize to unseen categories while maintaining robust performance on seen categories.

\subsection{General Image Classification}
\subsubsection{Experimental Setup}
\textit{Dataset.} We conduct general image classification experiments on the CIFAR10~\citep{krizhevsky2009learning} dataset. The CIFAR10 dataset consists of 60,000 images with a 5-1 training-testing split, divided into 10 mutually exclusive classes.

\noindent\textit{Implementation details.} In our experiments, we use WideResNet~\citep{zagoruyko2016wide} as the feature extractor. Similarly, features extracted from the feature
extractor are considered to be located in the tangent
space at the origin and then projected to the hyperbolic space via an exponential map added on the top
of the backbone. Assuming the dataset $\mathcal{D}$ consists of $n$ classes, we design a distance-based $(n+m)$-class classifier (containing $n$ original seen classes and $m$ unseen classes synthesized by HDFA) to perform classification in the hyperbolic space. Then, we use HDFA to perform augmentation for the training data.

\subsubsection{Results}
Table~\ref{tab:gen} shows the accuracy results on the general image classification experiment using
the CIFAR10 dataset. We observe that our augmentation method can improve the classification performance of the hyperbolic baseline network even on a dataset that lacks explicit hierarchical structure.

\subsection{Ablation}
We conduct ablation experiments in the class-incremental learning task on the Tiny-ImageNet dataset, and few-shot learning task on the mini-ImageNet and tiered-ImageNet datasets. `w/o Aug' denotes not performing any augmentation, where we directly use the original given data to train the classifier. `Inf Aug' denotes performing infinite augmentation only for seen classes. Specifically, we first do not use neural ODEs to estimate distributions of the seen classes, but directly compute the mean and covariance matrix from the given data and manually tune the curvature. Then, we use the neural ODEs to estimate the distributions for seen classes, and only generate 5 samples for each class, denoted by `Neural ODE (seen) + Aug 5 samples'. Next, we use the neural ODEs to estimate the distributions for seen classes and perform infinite augmentation, denoted by `HFA'. Finally, we perform our HDFA method, that is using the neural ODEs to estimate the distributions and performing infinite augmentation for both seen and unseen classes. Results are shown in Table~\ref{tab:final_tiny_abla} and Table~\ref{tab:abaltion}, showing the effectiveness of our dual augmentation, neural ODE, and the derived upper
bound of the augmentation loss.

More comparisons between our HDFA with HFA in the class-incremental learning, few-shot open-set recognition, few-shot learning, zero-shot learning, and general image classification tasks are shown in previous subsections, including Table~\ref{tab:final_cifar100}, Table~\ref{tab:final_tiny}, Table~\ref{tab:average_cifar100}, Table~\ref{tab:fsor}, Table~\ref{tab:fsor_c}, Table~\ref{tab:few_FC100}, Table~\ref{tab:few_meta}, and Table~\ref{tab:zero}. Our superior performance suggests that augmentation for both seen and unseen classes can make a better approximation to the real-world distribution, enhancing the model in both closed- and open-environment scenarios.

\begin{table}[t]
    \centering
    \caption{Training time (seconds) and memory usage (MB) comparisons on the 5-way 1-shot few-shot learning task using the mini-ImageNet dataset. All methods use ResNet12 as the backbone. }
    \begin{tabular}{lcc}
        \hline
         Methods & Time &  Memory \\
         \toprule
         \specialrule{0em} {1.5pt}{1.5pt}
         HyperProto~\citep{khrulkov2020hyperbolic}& 9.67$\times 10^{0}$ & 7.86$\times 10^{2}$  \\
         \specialrule{0em} {1.5pt}{1.5pt}
         HFA~\citep{gao2022hyperbolic}& 4.01$\times 10^{1}$& 1.58$\times 10^{3}$\\
         \specialrule{0em} {1.5pt}{1.5pt}
         \textbf{HDFA (Ours)}& 7.46$\times 10^{1}$& 2.62$\times 10^{3}$\\
         \hline
    \end{tabular}
    \label{tab:eff-mini}
\end{table}
\begin{table}[t]
    \centering
    \caption{Training time (seconds) and memory usage (MB) comparisons on the 5-way 1-shot few-shot open-set recognition task using the CIFAR-FS dataset. All methods use ResNet12 as the backbone.}
    \begin{tabular}{lcc}
        \hline
         Methods & Time &  Memory \\
         \toprule
         \specialrule{0em} {1.5pt}{1.5pt}
         ProtoNet~\citep{snell2017prototypical}& 5.51$\times 10^{2}$ & 3.36$\times 10^{3}$  \\
         \specialrule{0em} {1.5pt}{1.5pt}
         HFA~\citep{gao2022hyperbolic}& 5.81$\times 10^{2}$ & 3.50$\times 10^{3}$\\
         \specialrule{0em} {1.5pt}{1.5pt}
         \textbf{HDFA (Ours)}& 1.08$\times 10^{3}$ & 4.00$\times 10^{3}$ \\
         \hline
    \end{tabular}
    \label{tab:eff-cfs}
\end{table}

\begin{table}[t]
    \centering
    \caption{Training time (seconds) and memory usage (MB) comparisons on the class-incremental learning task in the C100-B40-S20 setting using the CIFAR100 dataset. All methods use ResNet18 as the backbone.}
    \begin{tabular}{lcc}
        \hline
         Methods & Time &  Memory \\
         \toprule
         \specialrule{0em} {1.5pt}{1.5pt}
         IL2A~\citep{zhu2021class}& 3.02$\times 10^{0}$ & 1.33$\times 10^{4}$  \\
         \specialrule{0em} {1.5pt}{1.5pt}
         HFA~\citep{gao2022hyperbolic}& 1.32$\times 10^{1}$ & 1.34$\times 10^{4}$\\
         \specialrule{0em} {1.5pt}{1.5pt}
         \textbf{HDFA (Ours)}& 3.46$\times 10^{1}$ & 1.74$\times 10^{4}$ \\
         \hline
    \end{tabular}
    \label{tab:eff-c100}
\end{table}

\subsection{Efficiency Analysis}
In this section, we provide a comprehensive evaluation of the computational efficiency of our method. Specifically, we evaluate the training time of each outer loop iteration and the overall memory footprint. We compare the efficiency of our method with the baseline methods and HFA across three tasks: few-shot learning, few-shot open-set recognition and class-incremental learning.  The results are shown in Table~\ref{tab:eff-mini}, Table~\ref{tab:eff-cfs} and Table~\ref{tab:eff-c100}, respectively. As demonstrated in the tables, our method shows comparable efficiency to both the baseline methods and HFA in terms of computational time and memory consumption. This indicates that our method can enhance the model performance without introducing substantial inefficiencies.

\begin{figure}[t]
    \centering
    \begin{subfigure}[b]{0.49\columnwidth}
        \centering
        \includegraphics[width=1\columnwidth]{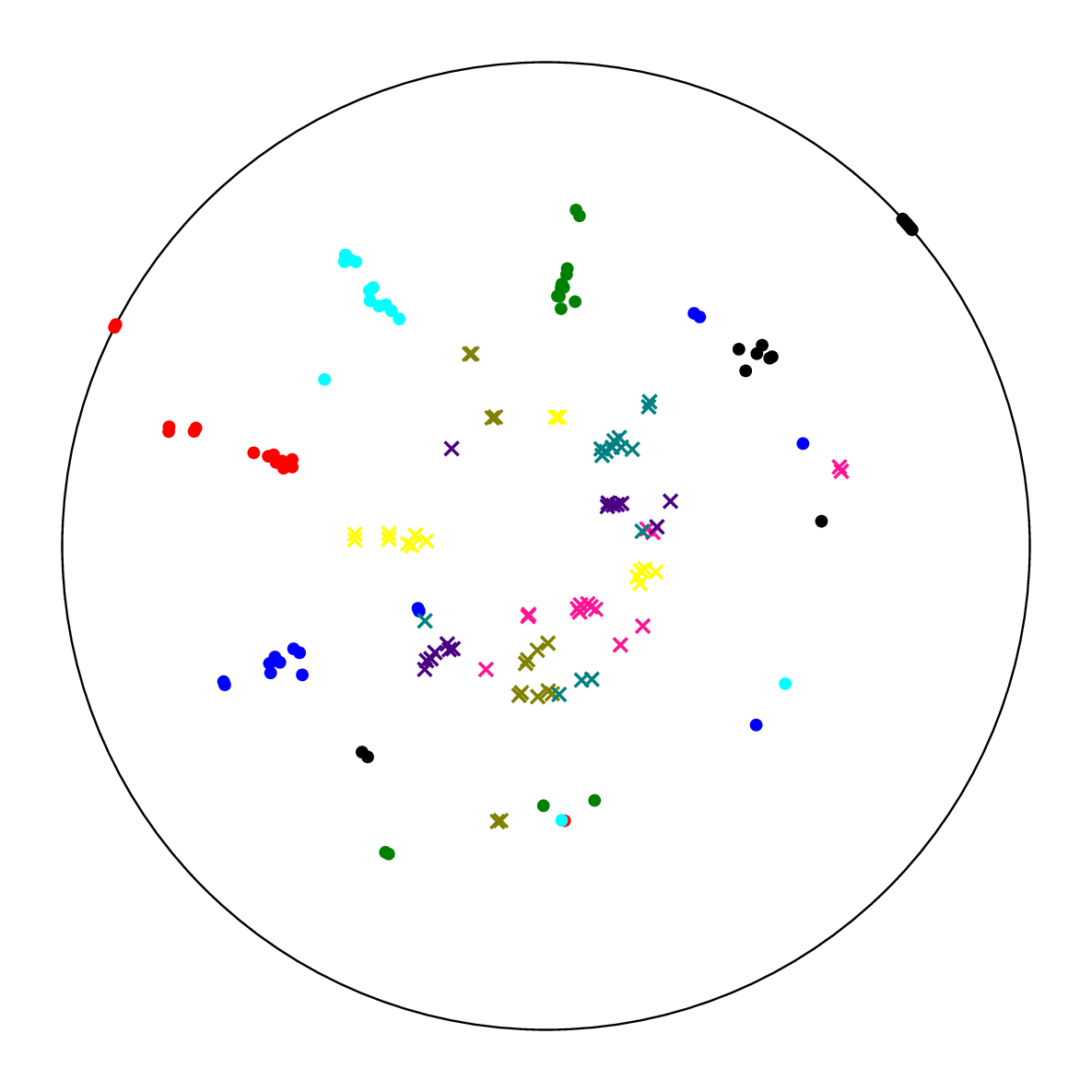}
        \caption{Features in the mini-ImageNet dataset.}
        \label{cosne}
    \end{subfigure}
        \begin{subfigure}[b]{0.49\columnwidth}
        \centering
        \includegraphics[width=1\columnwidth]{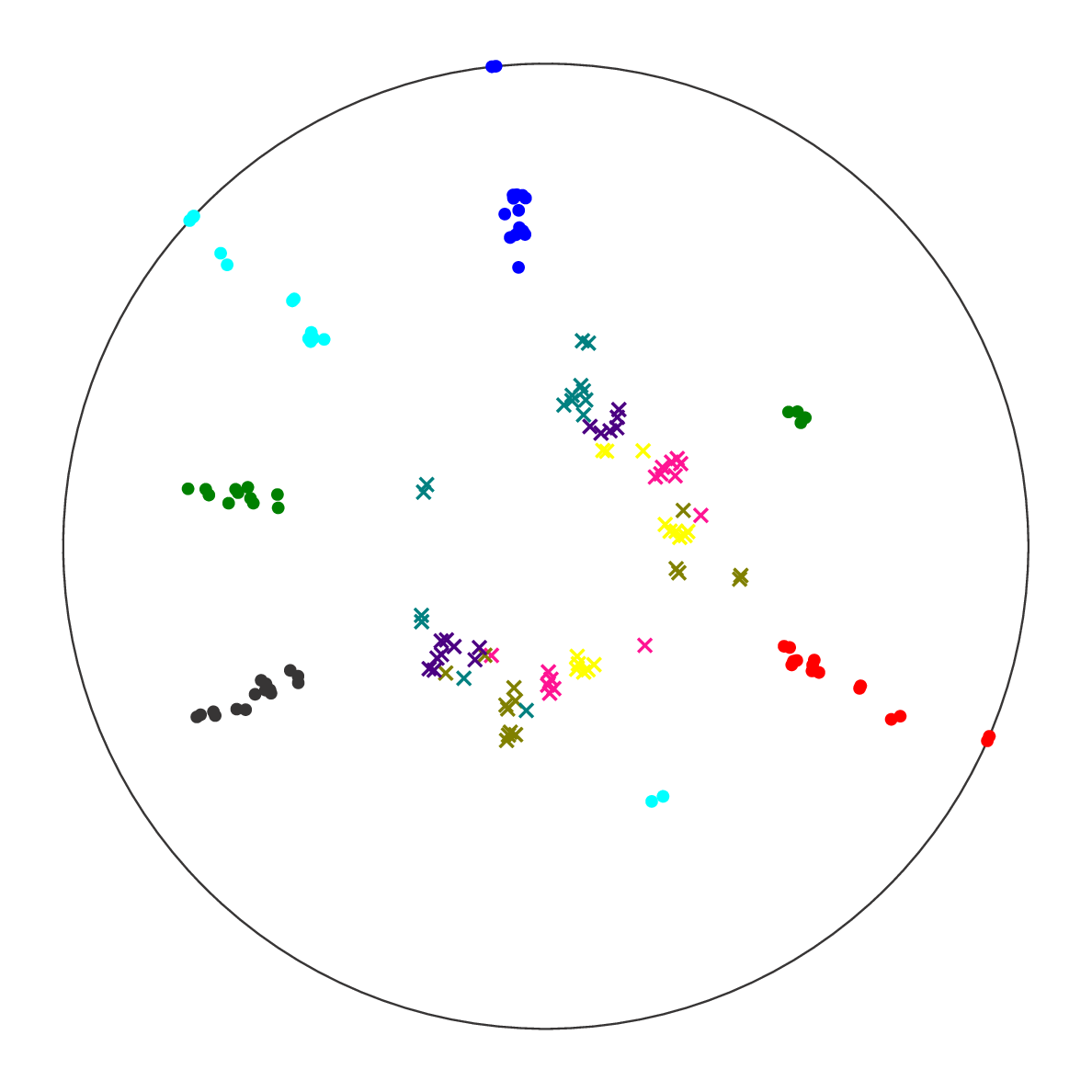}
        \caption{Features in the CUB dataset.\quad\quad}
        \label{tsne}
    \end{subfigure}
    \setlength{\abovecaptionskip}{0.cm}
    \begin{subfigure}{0.9\columnwidth}
    \includegraphics[width=0.95\columnwidth]{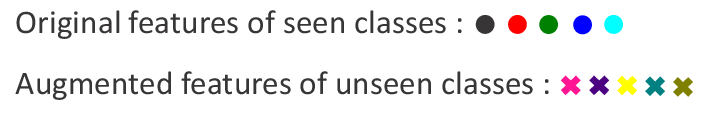}
    \end{subfigure}
    \caption{Visualization of augmentation features via CO-SNE in the mini-ImageNet and CUB datasets. The red, black, blue, green, and cyan dots denote features sampled from the distribu-
    tions of seen classes, and the x-shaped points in other colors denote features sampled from the distributions of unseen classes. The augmented features of unseen classes are closer to the center of the Poincar\'e ball with higher uncertainty.}
    \label{feature visual cosne}

\end{figure}
\begin{figure*}[t]
    \centering
    \begin{subfigure}[b]{0.65\columnwidth}
        \centering
        \includegraphics[width=0.7\columnwidth]{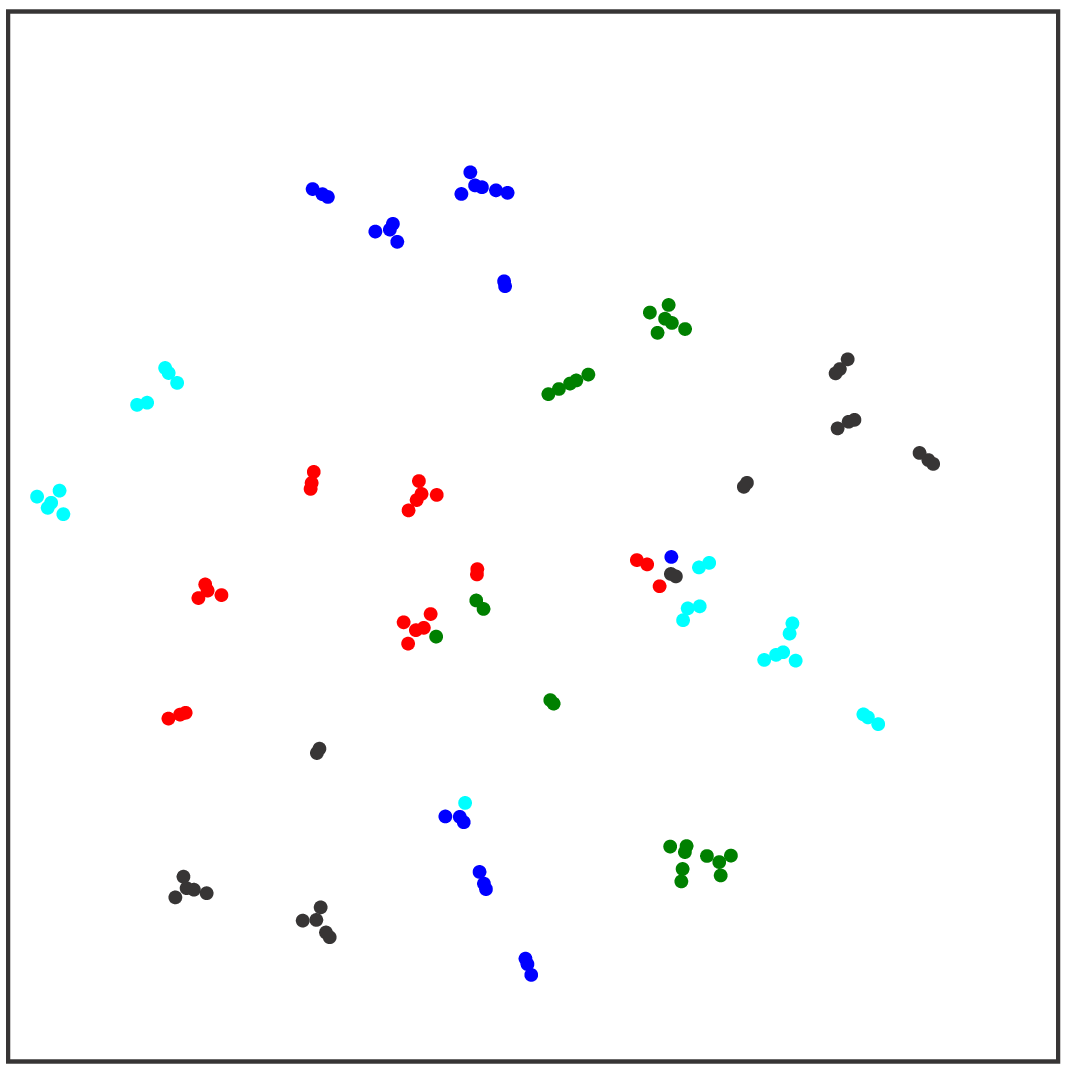}
        \caption{Original features in mini-ImageNet.}
        \label{cosne}
    \end{subfigure}
    \begin{subfigure}[b]{0.65\columnwidth}
        \centering
        \includegraphics[width=0.7\columnwidth]{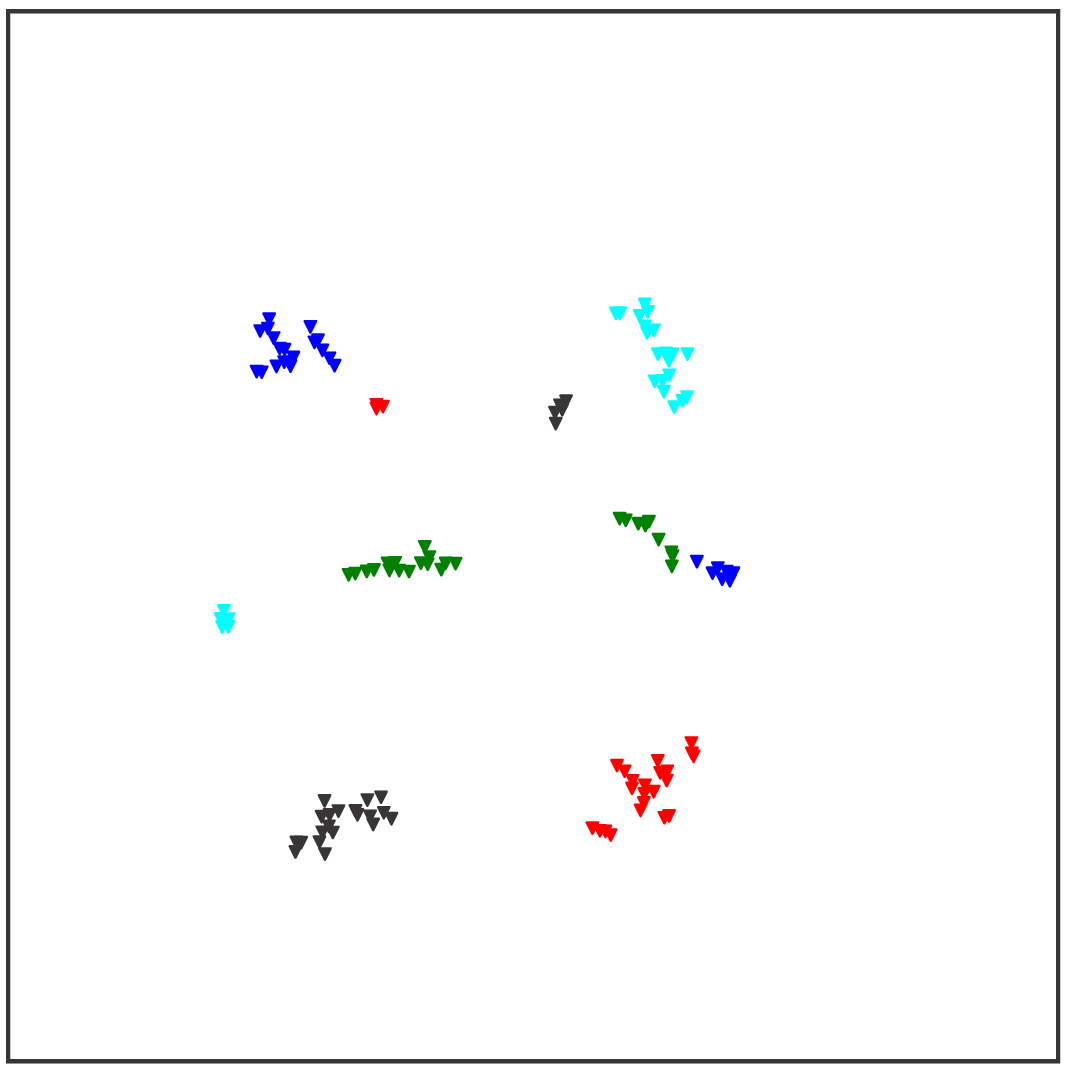}
        \caption{Features by HFA in mini-ImageNet.}
        \label{tnse}
    \end{subfigure}
        \begin{subfigure}[b]{0.65\columnwidth}
        \centering
        \includegraphics[width=0.7\columnwidth]{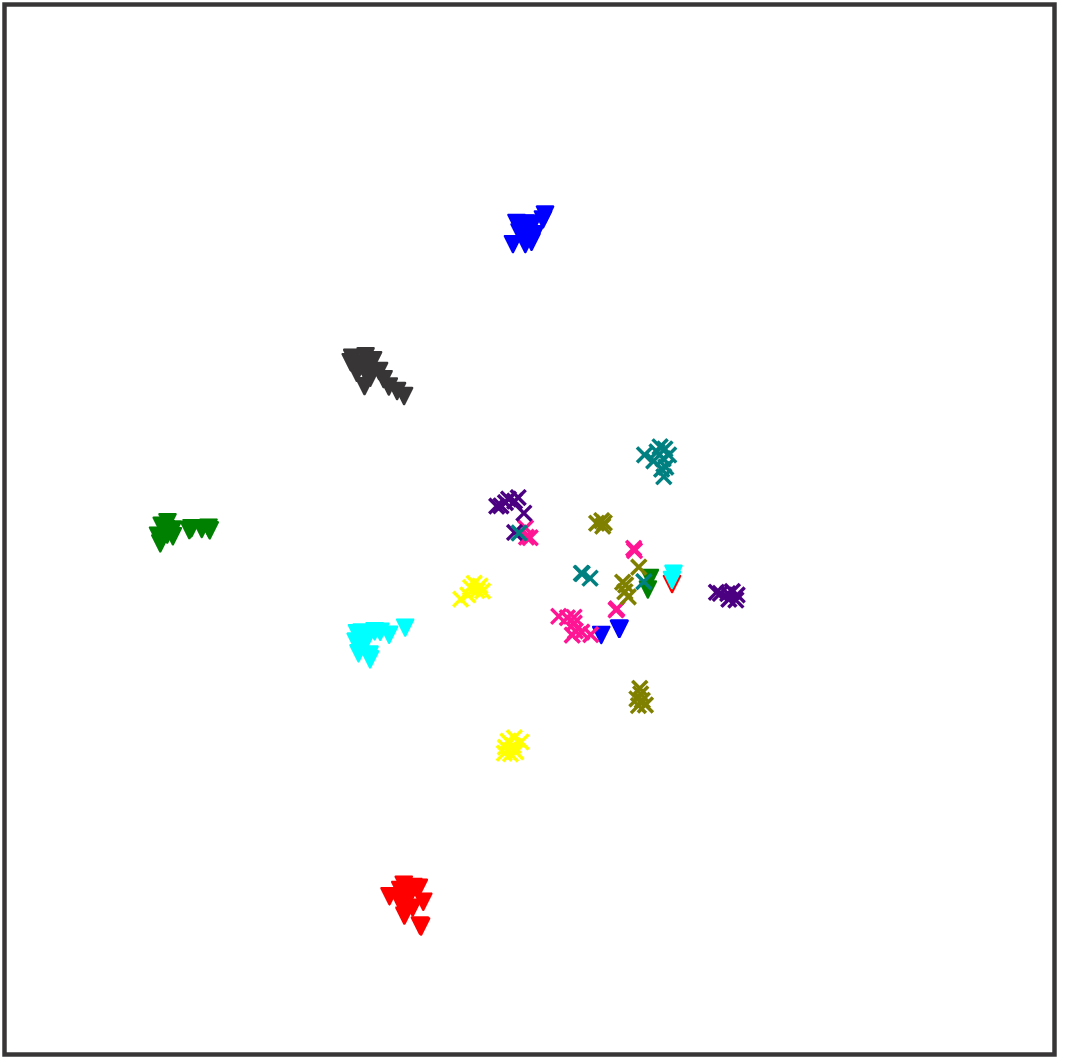}
        \caption{Features by HDFA in mini-ImageNet.}
        \label{tsne} 
    \end{subfigure}
    \begin{subfigure}[b]{0.65\columnwidth}
        \centering
        \includegraphics[width=0.7\columnwidth]{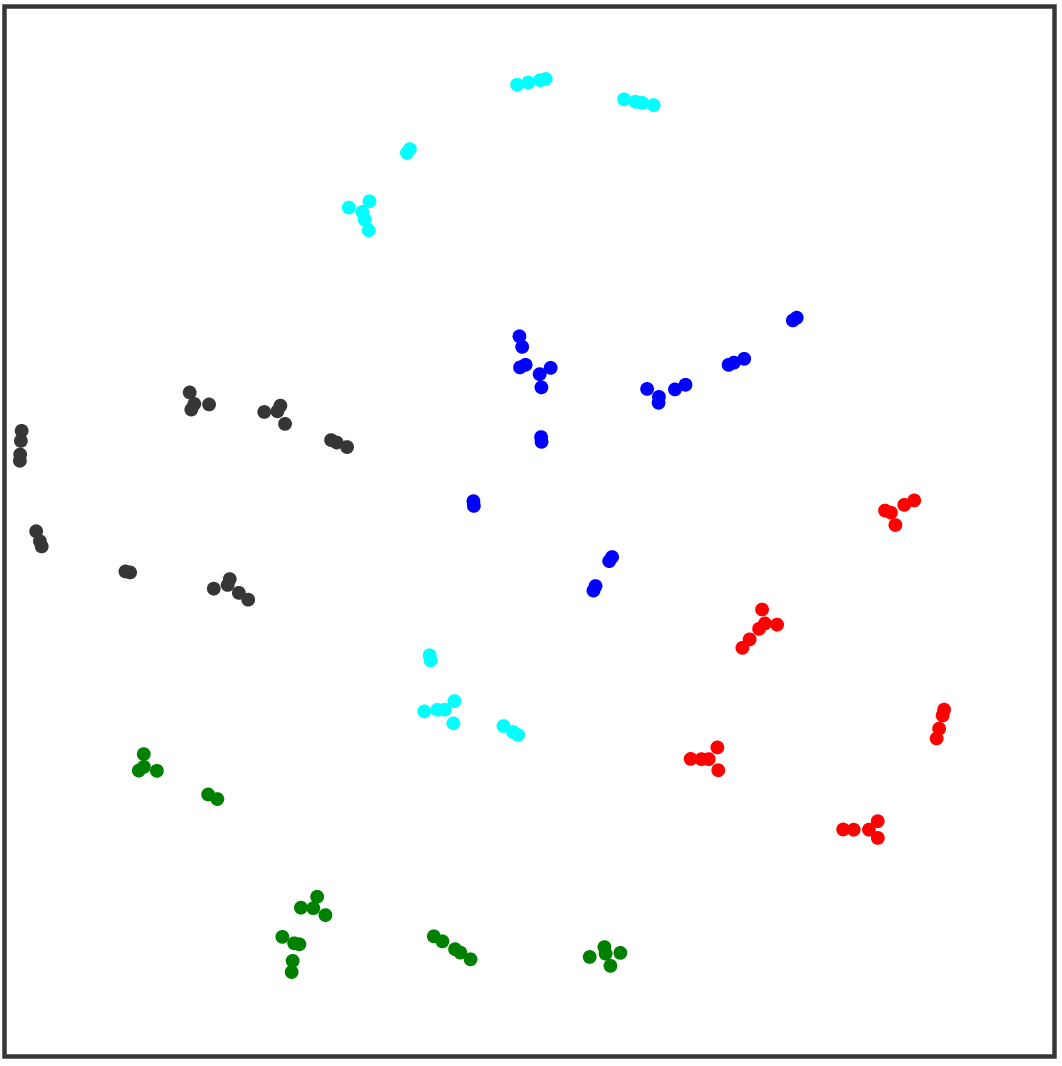}
        \caption{Original features in CUB.}
        \label{cosne}
    \end{subfigure}
    \begin{subfigure}[b]{0.65\columnwidth}
        \centering
        \includegraphics[width=0.7\columnwidth]{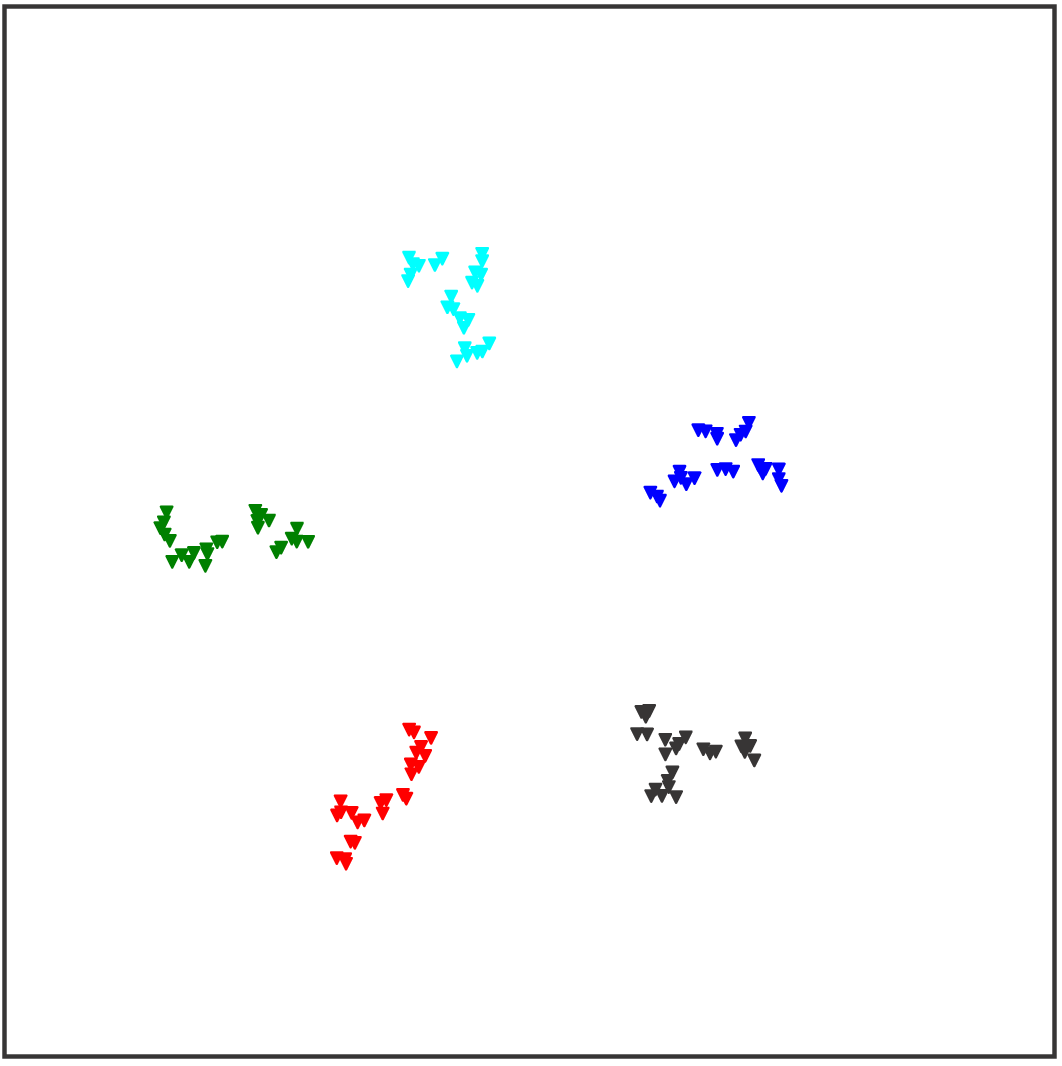}
        \caption{Features by HFA in CUB.}
        \label{tnse}
    \end{subfigure}
        \begin{subfigure}[b]{0.65\columnwidth}
        \centering
        \includegraphics[width=0.7\columnwidth]{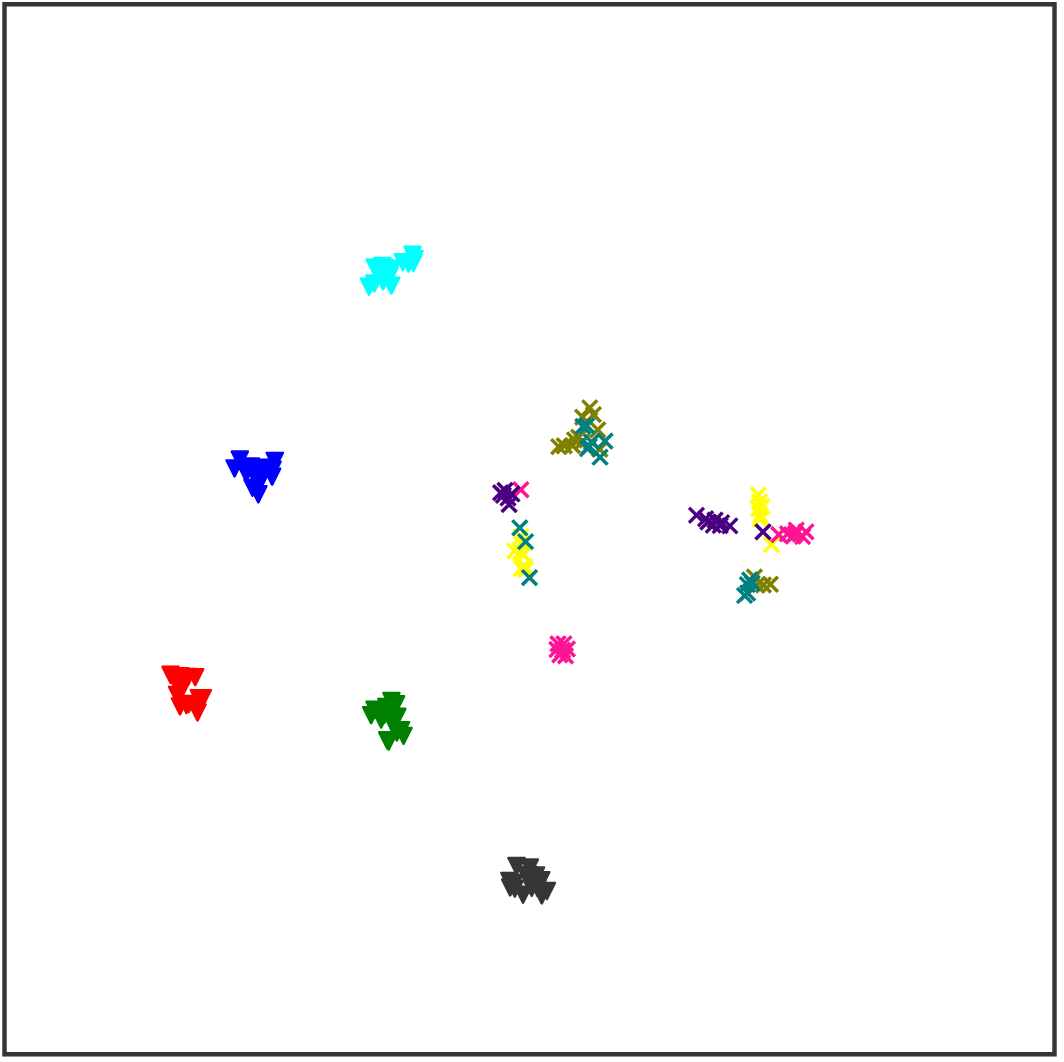}
        \caption{Features by HDFA in CUB.}
        \label{tsne}
    \end{subfigure}
    \setlength{\abovecaptionskip}{0.cm}
    \begin{subfigure}{2\columnwidth}
    \includegraphics[width=0.94\columnwidth]{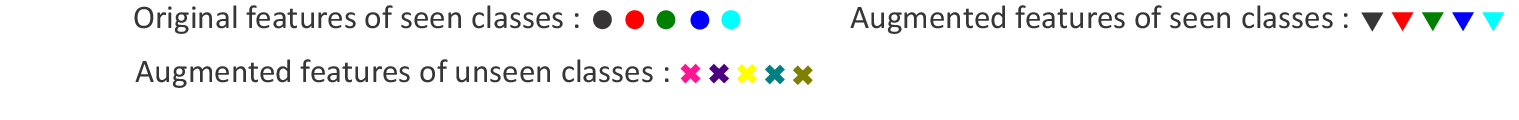}
    \end{subfigure}

    \caption{Visualization of augmentation features. The red, black, blue, green, and cyan dots denote the sampled features of 5 seen classes in the original dataset. The red, black, blue, green, and cyan triangles denote features sampled from the estimated distributions of seen classes, and the x-shaped points in other colors denote features sampled from the distributions of synthesized unseen classes. We observe that our method effectively alleviates the over-dominance issue of seen classes in the original feature space, leading to sharper classification boundaries, smaller intra-class distances, and larger inter-class distances.}
    \label{feature visual tsne}

\end{figure*}

\begin{figure*}[t]
\centering
\includegraphics[width=2\columnwidth]{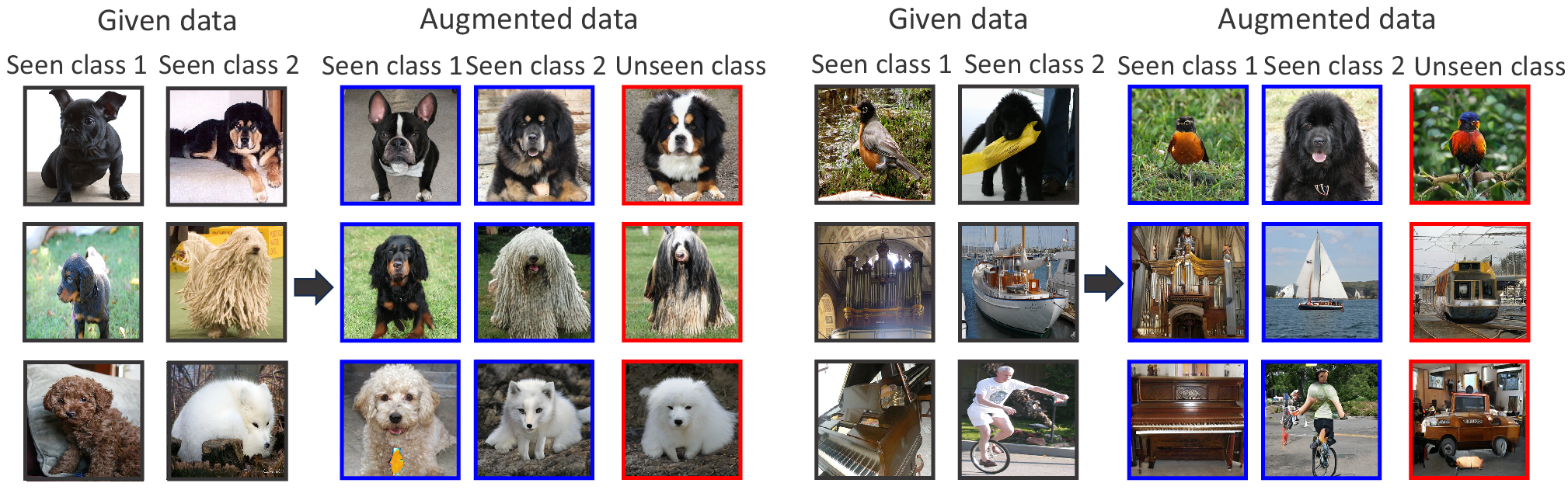}
\caption{Given data and augmented data by HDFA in the mini-ImageNet dataset.}
\label{aug-fig}
\end{figure*}

\subsection{Visualization}
In this section, we demonstrate the effect of HDFA with visualization results in the hyperbolic space.
We use the dimensionality reduction method: CO-SNE~\citep{guo2022co} 
to plot the estimated distribution in the mini-ImageNet and CUB datasets by sampling augmented features. Results are shown in Figure~\ref{feature visual cosne} and Figure~\ref{feature visual tsne}. In Figure~\ref{feature visual cosne}, we observe that augmented features of synthesized unseen classes are closer to the center of the Poincar\'e ball, and augmented features of seen classes are closer to the boundary. This is consistent with the fact that synthesized unseen classes have larger uncertainty.
As shown in Figure~\ref{feature visual tsne}, we observe that the proposed HDFA method leads to larger inter-class distances and smaller intra-class distances by generating features for both seen and unseen classes. In addition, introducing auxiliary unseen classes during training allows the model to better approximate the real distribution when the training data is biased, alleviating the over-dominance issue of seen classes and resulting in sharper and more robust classification boundaries among both seen and unseen classes.

Then, we use a reverse mapping algorithm~\citep{Wang2019ImplicitSD} to search images corresponding to the augmented features of seen and unseen classes in the mini-ImageNet dataset for intuitive visualization. Results are shown in Figure~\ref{aug-fig}. The first and second columns represent the original given images from two seen classes. The third and fourth columns represent the augmented images of two seen classes, and the fifth column represents the augmented images of the synthesized unseen classes. We observe that the images of seen and unseen classes generated by HDFA are meaningful and have diversity and plausibility.

\section{Conclusion}
\label{conclusion}
In this paper, we have presented a hyperbolic dual feature augmentation method for open-environment that generates diverse
and plausible features of seen and unseen classes. Our method introduces unseen classes outside the original distributions during model training, which is beneficial for alleviating the over-dominance issue and enhancing new class generalization in applications of open-environment. The distributions estimated by the introduced neural ODEs precisely approximate the real distributions in scarce data regimes, avoiding undesirable data distortion. Our proposed hierarchy-preserving regularizer can also preserve the latent hierarchical data structures. The derived upper bound of the dual augmentation loss enables an efficient augmentation algorithm in the hyperbolic space, improving model generalization and reducing computational loads. 
Comprehensive experimental results demonstrated the superiority of the proposed HDFA method in open-environment scenarios, which significantly improves the performance of hyperbolic algorithms in scarce data regimes. 

In this work, we assume that data has a uniform hierarchical structure and employs a single hyperbolic space for each task. However, real-world data may exhibit complex hierarchical structures with diverse local structures. As observed in the class-incremental learning experiments, the method using product manifolds to embed data outperforms our approach in certain contexts. In the future, we plan to explore data augmentation in product manifolds to tackle such complex data in real-world applications.

\noindent \textbf{Data Availibility } All datasets used in this study are open access and have been cited in the paper.

\bibliography{sn-bibliography}

\end{document}